\documentclass[]{interact}
\usepackage{multirow}

\usepackage{epstopdf}
\usepackage[table]{xcolor}

\usepackage{rotating}  
\usepackage{pdflscape}  

\usepackage{subfigure}

\usepackage[numbers,sort&compress,merge]{natbib}
\bibpunct[, ]{(}{)}{,}{n}{,}{,}

\theoremstyle{plain}

\theoremstyle{definition}

\theoremstyle{remark}

\usepackage{amsmath,amssymb,amsfonts}
\usepackage{algorithmic}
\usepackage{graphicx}
\usepackage{textcomp}
\usepackage{xcolor}
\usepackage{lmodern}
\usepackage{textcomp}
\def\BibTeX{{\rm B\kern-.05em{\sc i\kern-.025em b}\kern-.08em
    T\kern-.1667em\lower.7ex\hbox{E}\kern-.125emX}}
\begin{document}


\title{Data-Driven Heat Pump Management: Combining Machine Learning with Anomaly Detection for Residential Hot Water Systems}

\author{
\name{Manal Rahal\textsuperscript{a,b}\thanks{CONTACT Manal Rahal. Email: manal.rahal@kau.se}, Bestoun S. Ahmed\textsuperscript{a,c}, Roger Renström\textsuperscript{a}, Robert Stener\textsuperscript{b}, and Albrecht Wurtz\textsuperscript{b}}
\affil{\textsuperscript{a}Department of Mathematics and Computer Science, Karlstad University, Karlstad, Sweden; 
\textsuperscript{b}Department of Research and Development, Thermia AB, Arvika, Sweden; \\
\textsuperscript{c}American Univesity of Bahrain, Riffa, Bahrain. 
}
}

\maketitle

\begin{abstract}

Heat pumps (HPs) have emerged as a cost-effective and clean technology for sustainable energy systems, but their efficiency in producing hot water remains restricted by conventional threshold-based control methods. Although machine learning (ML) has been successfully implemented for various HP applications, optimization of household hot water demand forecasting remains understudied. This paper addresses this problem by introducing a novel approach that combines predictive ML with anomaly detection to create adaptive hot water production strategies based on household-specific consumption patterns. Our key contributions include: (1) a composite approach combining ML and isolation forest (iForest) to forecast household demand for hot water and steer responsive HP operations; (2) multi-step feature selection with advanced time-series analysis to capture complex usage patterns; (3) application and tuning of three ML models: Light Gradient Boosting Machine (LightGBM), Long Short-Term Memory (LSTM), and Bi-directional LSTM with the self-attention mechanism on data from different types of real HP installations; and (4) experimental validation on six real household installations. Our experiments show that the best-performing model LightGBM achieves superior performance, with RMSE improvements of up to 9.37\% compared to LSTM variants with $R^2$ values between 0.748-0.983. For anomaly detection, our iForest implementation achieved an F1-score of 0.87 with a false alarm rate of only 5.2\%, demonstrating strong generalization capabilities across different household types and consumption patterns, making it suitable for real-world HP deployments.

\end{abstract}


\section{Introduction}
Heat pumps (HPs) have emerged as a vital component in supporting the global goal of low-carbon heating, positioning them among the most promising technologies. An HP is a device that transfers energy from low-temperature heat sources, such as outdoor air or water, to efficiently provide heating and hot water in residential and commercial buildings \cite{Gaur2021}. The availability of hot water that meets household needs must be sufficient at any time and in any season, while ensuring production efficiency. The speed and responsiveness of the HP are crucial to ensure user comfort and prevent cold showers \cite{Sadjjadi2023}.
Energy forecasting plays an integral part in modern energy management systems, more importantly in the transition towards greener energy \cite{Forootan2022}. Energy forecasting, particularly in predicting electricity demand, is widely applied at all levels, including generation, transportation, and distribution \cite{AhmadT2020}. The accurate forecasting of energy demand allows better resource allocation, reduces operational costs, and supports efficient integration of renewable resources \cite{Ukoba2024}.

However, forecasting household demand for hot water in an HP is one of those applications that has rarely been studied and remains under-explored in the literature. In the specific case of HPs, demand forecasting is crucial to ensure household comfort and enhance the efficiency of the broader energy grid. Forecasting household demand for hot water production in an HP requires understanding and analyzing household behavior with respect to tap water consumption and identifying key predictors. This is enabled through smart meters and temperature sensors strategically installed in the HP and its corresponding water tank. These sensors provide detailed insights into historical consumption patterns, allowing effective demand forecasting tasks.

This focus on energy efficiency and forecasting becomes particularly significant when considering that around 80\% of the greenhouse gas emissions are due to energy demand in the USA and the European Union\footnote[1]{https://world-nuclear.org/information-library/energy-and-the-environment/carbon-dioxide-emissions-from-electricity}. In this context, HPs have emerged as a cost-effective and clean energy alternative due to their low-carbon heating technology \cite{Rosenow2022}. The widespread adoption of HPs in recent decades reflects their vital role in both energy efficiency and sustainable development, offering significant reductions in carbon emissions and supporting the transition towards sustainable energy \cite{Gaur2021}. According to data from the International Energy Agency (IEA), around 600 million HPs will be installed globally by 2030 \cite{Song2023consumption}. Therefore, investing in efficient and smarter HPs is in everyone's interest.

Several approaches have been explored for energy management and forecasting in the literature. Traditional forecasting methods, such as autoregressive models, autoregressive integrated moving average, and exponential smoothing, are often used in load and price forecasting \cite{Antonopoulos2020}. However, as highlighted by \cite{Antonopoulos2020}, real-world demand response systems incorporate sensors that exhibit non-linear behaviors, making traditional linear modeling approaches insufficient for accurate forecasting.

Due to the global focus on the development of efficient HPs, there has been significant progress in the development of smart sensors that enable low-cost computational power \cite{Song2023}. Various sensors are being integrated into HPs, producing large amounts of data. Data collected from smart sensors are essential for performing data-driven analytics to understand and forecast demand trends \cite{YilDiz2017}. The availability of such high-frequency data enables vast applications in advanced analytics and performance optimization supported by machine learning (ML) systems. Predicting future energy demand allows for better decision-making, effective energy planning, and better customer comfort \cite{Bayram2023}.

Despite these advances, several limitations exist in current approaches. Heating, ventilation, and air conditioning systems (HVAC), among other engineering systems, have adopted ML techniques such as supervised and unsupervised learning, computer vision, neural networks (NNs) for demand forecasting, fault detection, and evaluation of customer behavior \cite{Zhang2023}. However, the millions of entries collected from HPs are rarely investigated, leading to missed opportunities for valuable insights \cite{Song2023}. The limited utilization of HP's data could be attributed to the complexity of analyzing high-granularity data, the low quality of the collected data, and the lack of knowledge of innovative use cases. In practice, the quality of the HP-generated data will probably suffer from missing values, inaccuracy, and inconsistency of sensor recordings \cite{Song2023}. As a result, the full potential of data-driven HP systems remains untapped. While NNs excel in capturing the complex, non-linear relationships in the data, leading to better forecasting performance \cite{Antonopoulos2020}, their application to household hot water demand forecasting remains limited.

The number of studies in modeling data collected from a real-world HP installation is limited to a specific range of applications. Most research investigates performance optimization and fault detection issues in large-scale installations with many connected users \cite{Song2023consumption}. To address this gap, this paper investigates the application of ML to predict user demand for hot water in households of real HP installations, where "hot water" refers to hot tap water throughout this paper. The main aim of this paper is to utilize ML and anomaly detection methods to develop demand-responsive HP. This is achieved by adapting the hot water function in the HP to the demand of the household since the demand trend is influenced by factors such as seasons, holiday months, and changes in family size. As a result, we propose a composite ML-anomaly detection approach that implements the following tasks:

\begin{enumerate}
\item A novel composite approach combining ML and isolation forest (iForest) is proposed to forecast household demand for hot water and steer responsive HP operations. Unlike existing methods that use either ML for prediction or anomaly detection separately, our approach integrates both techniques to identify consumption events, creates household-specific calendars, and enables precise steering based on predicted demand rather than fixed thresholds.

\item A systematic multi-step feature selection approach is investigated to identify important raw and lag input variables reducing the computational complexity of ML models and enhancing their performance. This addresses limitations in existing approaches that often use predetermined features, by employing statistical testing to identify optimal variables, reducing computational complexity by up to 68\%, and effectively capturing complex temporal dependencies.

\item Comparative analysis and tuning of three ML models, light gradient boosting machine (lightGBM), long short-term memory (LSTM), bi-directional LSTM with self-attention mechanism (attLSTM) on data collected from different types of real HP installations. Our approach differs from existing work by testing both ensemble and deep learning models on real-world data, providing household-specific tuning strategies, and demonstrating that LightGBM achieves superior performance for this application.

\item To demonstrate the effectiveness of the proposed approach, experiments are designed to test it on six household data with real HP installations. Our results show performance improvements of up to 9.37\% in RMSE compared to existing methods, successful anomaly detection with F1-score of 0.87, and practical implementation insights for residential settings.

\end{enumerate}

The novelty of our approach lies in the integration of multiple ML models, namely LightGBM, LSTM, BiLSTM with attention (attLSTM), and iForest, to provide a comprehensive framework for hot water consumption prediction and anomaly detection in household heat pump data. Unlike previous works that primarily focus on individual models, our study demonstrates the complementary strengths of each technique. LightGBM provides a lightweight and efficient baseline model, while LSTM and BiLSTM-attention are designed to capture complex temporal patterns. Additionally, iForest enhances robustness by identifying anomalous shower events effectively. The proposed combination offers improved prediction accuracy and robustness compared to conventional approaches. Furthermore, our framework is optimized for practical deployment, making it a valuable contribution to time-series forecasting and anomaly detection in energy-related systems.

The remainder of this paper is structured as follows: Section \ref{sec:background} summarizes the relevant background concepts and lays out the terminology necessary to understand the paper. Section \ref{sec:approach} discusses the proposed approach. Section \ref{sec:methodology} introduces the methods and models used to implement our approach. Section \ref{sec:dataanalysis} presents the household datasets and discusses the generation of raw and time-related characteristics. The implementation and empirical evaluation are presented in Section \ref{sec:results} and discussed in Section \ref{sec:discussion}. In Section \ref{sec:relatedwork}, similar work is highlighted. Finally, the main conclusions are drawn in Section \ref{sec:conclusion}.

\section{Related work}\label{sec:relatedwork}

The literature on demand forecasting constitutes a large portion of the recent literature related to energy, if not the largest \cite{Hernandezmatheus2022}. Understanding consumer behavior patterns and appliance usage, such as HPs, has attracted interest, particularly in ensuring a user-friendly response to demand with a greater focus on residential settings \cite{Antonopoulos2020}. The main objectives are usually to minimize the electricity cost and wasted energy in the surroundings while ensuring the customer's comfort \cite{Antonopoulos2020}. ML is one of the efficient methods used to forecast power demand, optimize the use of energy assets, and better understand energy usage patterns, showing improved accuracy and shorter evaluation time compared to other techniques \cite{Antonopoulos2020, Aprea2017}. The forecasting accuracy of ML algorithms for time series data forecasting is better than that of regression models \cite{ADELEKAN2022}. These methods rely on learning consumption patterns from historical consumption data to predict user demand \cite{Somu2021}. In fact, the existing literature shows that supervised ML algorithms perform well in predicting future load demand, wind, solar, and geothermal energy forecasting in a wide range of HVAC applications \cite{AhmadChen2020}. Recently, deep learning has become popular in fault detection and diagnosis, particularly LSTM, because it excels at modeling long-term temporal dependencies \cite{Cabrera2020}. Although most literature focuses on forecasting energy demand \cite{Ahmad2020}, other applications exist but are less prevalent, including those related to HP. This is because accurately forecasting household demand is not a straightforward task \cite{Abbasimehr2020}. Inspired by the success of other energy-related applications, the HP industry has utilized ML to enhance the performance of HP systems, including modeling, prediction, and fault detection tasks. Without a doubt, ML methods perform very well in solving HVAC system-related problems due to the complicated and non-linear nature of such systems \cite{Mirnaghi2020}. However, most HP-related research focuses on the residential sector \cite{Schlosser2020}. 

However, several studies have used various ML models to predict energy consumption in office buildings. For example, \cite{Sendra2020} used artificial neural networks (ANN) to predict the heating load of residential buildings in Agadir, while \cite{Xypolytou2017} used ANN to accurately forecast the energy consumption of HP systems in office buildings. Similarly, \cite{Zhou2020} and \cite{Dong2016} tested ANN alongside other models to forecast the energy consumption of air conditioning systems.

LSTM models have also been widely applied. \cite{Sendra2020} used LSTM for one-day-ahead power consumption prediction of a building HVAC system, and \cite{Naug2018} used LSTM to forecast current energy consumption based on previous time step information. \cite{Kong2017} concluded that LSTM outperforms other models in short-term load forecasting of hot water systems in residential households. Furthermore, \cite{Wang2020} found that LSTM is highly accurate in predicting a building's thermal load, comparing its performance with models such as linear regression, support vector machines, random forests, and naïve models. Moreover, \cite{Zhou2020} and \cite{Dong2016} also applied LSTM to predict energy consumption in air conditioning systems.

The extreme learning machine models were compared by \cite{Huang2019} to support vector regression, multiple linear regression, and ensemble learning models to forecast a two-hour heating load in HP systems. Support vector machines were further examined by \cite{Wang2020} in predicting the thermal loads of buildings. The same study highlighted ensemble learning models, particularly the Extreme Gradient Boost, for their accuracy. In another application, \cite{DAS2022b} introduced a fuzzy-ELM hybrid approach that utilizes fuzzy inference systems to improve the forecasting accuracy of oil and gold price data under uncertainty. Integrating fuzzy inference systems with ELM demonstrated promising results, enhancing the model's ability to manage nonlinear data and improve overall performance.

While ELM has been successfully applied in various forecasting domains, their optimization attracted researchers' interest to further improve the performance. Recent work by \cite{Das2022} shows how parameter optimization using particle swarm optimization (PSO) and variants of the crow search algorithm improve ELM in forecasting stock prices \cite{Das2022}. Their results show that the optimized models outperform the existing algorithms in solving 12 benchmark problems. In addition, \cite{SAHU2021} applied modified teaching learning-based optimization to optimize ELM’s performance in high-dimensional power forecasting data. As ML applications in energy systems evolve, emerging technologies such as quantum artificial intelligence show promising potential for revolutionizing energy production forecasting \cite{baklaga2023}. While such advanced approaches are still emerging, they highlight the rapidly evolving landscape of ML applications in energy optimization.

Recent research has increasingly focused on integrating deep learning models with anomaly detection techniques for enhanced fault detection capabilities in various systems. Bouchenak et al. \cite{Bouchenak2022} proposed a semi-supervised deep learning approach that combines generative adversarial networks (GANs) with anomaly detection methods for modulation identification in MIMO systems. Their approach demonstrated the effectiveness of using deep learning for feature extraction coupled with anomaly detection to improve classification accuracy in complex signal environments.

In the energy domain specifically, Harrou et al. \cite{HARROU2024118665} introduced a semi-supervised framework that integrates variational autoencoders (VAE) with multiple anomaly detection algorithms, including iForest and One-Class SVM, for fault detection in grid-connected photovoltaic systems. Their approach achieved impressive accuracy rates of up to 92.90\% for MPPT mode and 92.99\% for IPPT mode in identifying various fault scenarios. The study highlights how combining deep learning's feature extraction capabilities with specialized anomaly detection methods can significantly enhance fault detection performance in energy systems while requiring only normal operation data for training.

These recent studies demonstrate a growing trend toward hybrid approaches that leverage both deep learning and traditional anomaly detection techniques to improve system reliability and efficiency across different domains. The integration of methods like iForest with deep learning models is particularly relevant to our work on heat pump systems, as it offers potential pathways for enhancing anomaly detection in residential energy consumption patterns.

Recent advances in transformer-based anomaly detection techniques are also relevant to this discussion. For instance, Haq et al. \cite{Haq2024} introduced TransNAS-TSAD, a framework that harnesses transformers for multi-objective neural architecture search in time series anomaly detection, which could be beneficial for optimizing anomaly detection in household heat pump data. Additionally, Lee et al. \cite{Lee2023} demonstrated the effectiveness of peak anomaly detection techniques from environmental sensor-generated time series data, which highlights the relevance of advanced anomaly detection methods for similar applications. These approaches provide valuable insights into the development of robust anomaly detection systems applicable to various domains.

Table~\ref{tab} provides a comprehensive comparison of previous studies on energy forecasting and anomaly detection in energy systems. The table categorizes approaches into statistical models, machine learning models, deep learning methods, and hybrid approaches, highlighting their applications, advantages, limitations, and performance metrics. This comparison illustrates the evolution of methods from traditional statistical approaches to sophisticated hybrid models that combine deep learning with anomaly detection techniques.

\begin{landscape}
\begin{table}
\centering
\caption{Comparison of Previous Studies on Energy Forecasting and Anomaly Detection}
\label{tab}
\begin{tabular}{|p{3.2cm}|p{3cm}|p{3cm}|p{3.5cm}|p{3.5cm}|p{3.5cm}|}
\hline
\textcolor{blue}{\textbf{Category}} & \textcolor{blue}{\textbf{Methods}} & \textcolor{blue}{\textbf{Studies}} & \textcolor{blue}{\textbf{Application}} & \textcolor{blue}{\textbf{Advantages}} & \textcolor{blue}{\textbf{Limitations}} \\
\hline
\textcolor{blue}{\textbf{Statistical \newline Models}} & \textcolor{blue}{ARIMA, SARIMA} & \textcolor{blue}{\cite{Zhou2020}, \cite{Dong2016}} & \textcolor{blue}{Energy consumption forecasting} & \textcolor{blue}{Simple implementation, works well with linear data} & \textcolor{blue}{Poor performance with non-linear patterns, requires stationary data} \\
\cline{2-6}
& \textcolor{blue}{Exponential Smoothing} & \textcolor{blue}{\cite{Antonopoulos2020}} & \textcolor{blue}{Load and price forecasting} & \textcolor{blue}{Handles seasonality, simple to implement} & \textcolor{blue}{Limited capacity for complex patterns} \\
\hline
\textcolor{blue}{\textbf{Machine \newline Learning \newline Models}} & \textcolor{blue}{ANN} & \textcolor{blue}{\cite{Sendra2020}, \cite{Xypolytou2017}} & \textcolor{blue}{Building energy consumption, HP systems} & \textcolor{blue}{Captures non-linear patterns} & \textcolor{blue}{Requires large datasets, prone to overfitting} \\
\cline{2-6}
& \textcolor{blue}{Support Vector Machines} & \textcolor{blue}{\cite{Wang2020}} & \textcolor{blue}{Thermal load prediction} & \textcolor{blue}{Works well with high-dimensional data} & \textcolor{blue}{Computationally expensive for large datasets} \\
\cline{2-6}
& \textcolor{blue}{Extreme Learning Machine} & \textcolor{blue}{\cite{Huang2019}, \cite{DAS2022b}} & \textcolor{blue}{HP heating load forecasting} & \textcolor{blue}{Fast training speed} & \textcolor{blue}{Sensitivity to parameter selection} \\
\hline
\textcolor{blue}{\textbf{Deep \newline Learning \newline Methods}} & \textcolor{blue}{LSTM} & \textcolor{blue}{\cite{Kong2017}, \cite{Wang2020}, \cite{Sendra2020}} & \textcolor{blue}{Hot water systems, thermal load prediction} & \textcolor{blue}{Captures temporal dependencies} & \textcolor{blue}{Requires extensive hyperparameter tuning} \\
\cline{2-6}
& \textcolor{blue}{Bi-LSTM with Attention} & \textcolor{blue}{Our work} & \textcolor{blue}{Hot water demand forecasting} & \textcolor{blue}{Enhanced feature extraction} & \textcolor{blue}{More complex architecture} \\
\hline
\textcolor{blue}{\textbf{Hybrid \newline Approaches}} & \textcolor{blue}{VAE + Isolation Forest} & \textcolor{blue}{\cite{HARROU2024118665}} & \textcolor{blue}{Fault detection in photovoltaic systems} & \textcolor{blue}{High accuracy (92.9\%), needs only normal data for training} & \textcolor{blue}{Complex implementation} \\
\cline{2-6}
& \textcolor{blue}{GAN + Anomaly Detection} & \textcolor{blue}{\cite{Bouchenak2022}} & \textcolor{blue}{Modulation identification} & \textcolor{blue}{Improved classification in complex environments} & \textcolor{blue}{Requires careful balancing during training} \\
\cline{2-6}
& \textcolor{blue}{LightGBM + Isolation Forest} & \textcolor{blue}{Our work} & \textcolor{blue}{Hot water demand forecasting, anomaly detection} & \textcolor{blue}{Efficient for large datasets, accurate shower event detection} & \textcolor{blue}{Requires household-specific tuning} \\
\hline
\end{tabular}
\end{table}
\end{landscape}

Furthermore, \cite{Zhou2020} and \cite{Dong2016} explored auto-regressive integrated moving averages, Gaussian process regression, and Gaussian mixture models for forecasting energy consumption in air conditioning systems. This comprehensive analysis shows the various approaches and the effectiveness of using different ML models in predicting energy consumption for various systems and settings.

\section{Background}\label{sec:background}


Heating is not the only function of an HP, it also offers cooling services to provide users with thermal comfort throughout different seasons. In other words, the main function of an HP is to provide heating during colder months and cooling during warmer months.

\subsection{Types of a heat pump}
There are three main types of HP: air-to-air, water source and geothermal \cite{SARBU2014}. The type of an HP is defined based on the input energy source. The heat source in an HP can be gas or air, surface water (river, lake, sea), groundwater, or soil \cite{SARBU2014}. The air source HP (ASHP) uses outside air as a medium for heat exchange. Air-to-air HPs are the least expensive and most common among all types of HP \cite{SARBU2014}. The water source HP (WSHP) uses water sources such as lakes, rivers, or groundwater as a source of heat. WSHPs are known to be highly efficient, but their use is limited to the availability of large water sources near deployment sites \cite{Gaur2021}. The third type is ground source HP (GSHP), which uses the natural heat energy stored in the ground as a heat source \cite{Gaur2021}. GSHP are widely used to provide thermal comfort in buildings due to their higher energy efficiency compared to other types \cite{ZHANG2022}. The popularity of GSHPs has increased due to their ability to integrate into smart energy systems and their reliable performance in extreme climates \cite{Ahmed2023}. Other types of HPs could include hybrid HP systems, which consist of traditional heating systems such as gas boilers or electric heaters connected to an HP \cite{SARBU2014}. In our use case, the data from six households is collected from two types of HPs, GSHP, and ASHP, where both types integrate heating and hot water production functions.

\subsection{Hot water demand forecasting problem}

Forecasting the demand for hot water production in an HP requires understanding household behavior regarding tap water consumption and identifying the key predictors. This is enabled by smart meters and temperature sensors strategically installed in the HP and its corresponding water tank to monitor and record relevant data. These recorded data are of a time series nature, providing detailed insights into historical consumption patterns and enabling demand forecasting tasks.
\begin{figure}
\centering
    \includegraphics[width=1\linewidth]{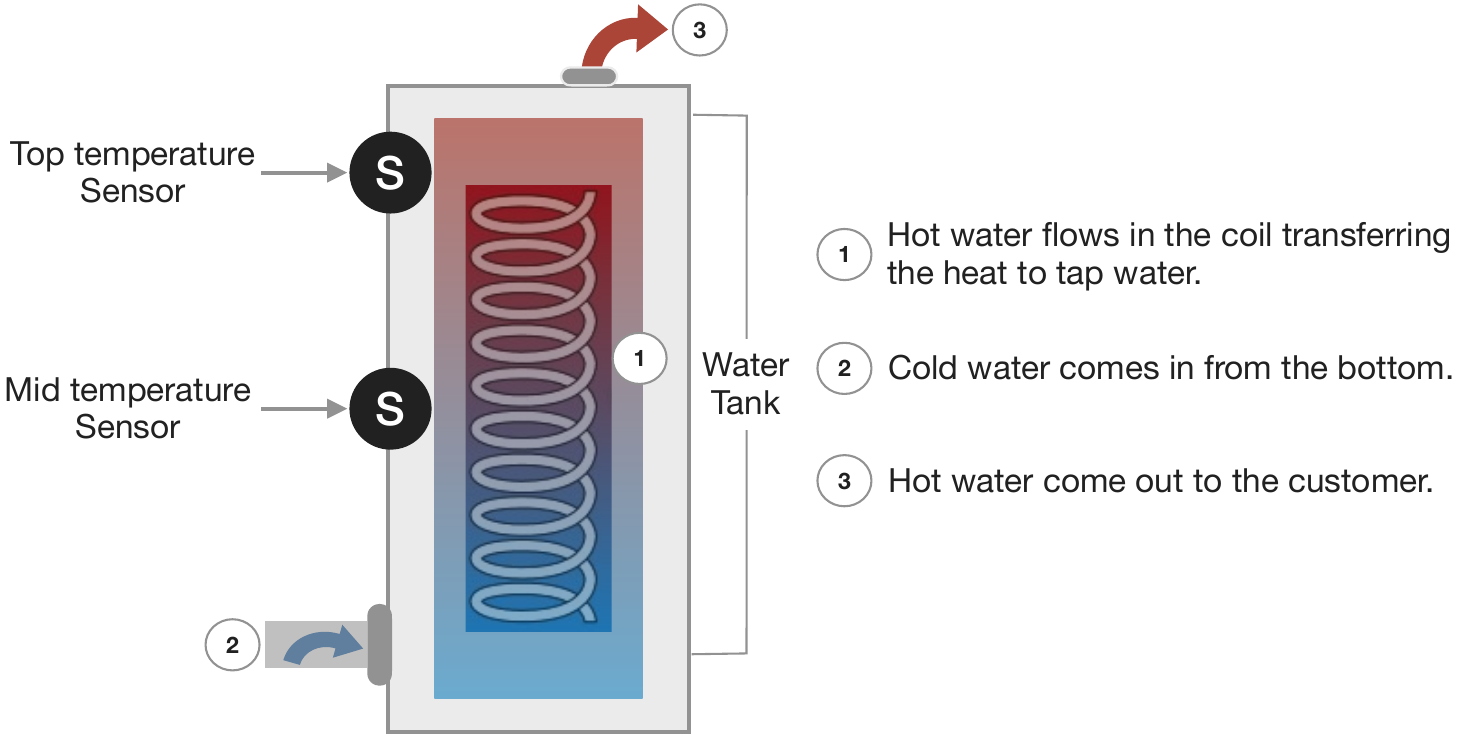}
    \caption{The HP water tank operations}
    \label{fig:tapwater}
\end{figure}
As observed in Figure \ref{fig:tapwater}, which illustrates the production of the hot water function, hot water circulates through the coil, transferring heat to the water in the tank. As the produced hot water accumulates at the top of the tank, it is used by the user for various household activities such as shower and dishwashing. As the water tank is replenished with cold water that enters from the bottom, the top sensor detects the temperature at the top of the tank ($t_\mathrm{top}$), and the middle sensor detects the temperature at the middle of the tank ($t_\mathrm{mid}$). During large consumption of hot water, $t_\mathrm{mid}$ drops significantly, while $t_\mathrm{top}$ maintains stable temperature levels. The availability of hot water at the top of the tank ensures a comfortable experience for the user.

The HPs included in this study start and stop hot water production based on preset thresholds for the average temperature between $t_\mathrm{mid}$ and top $t_\mathrm{top}$. As a result of the existing system, production may start when there is low or no demand for hot water due to the natural decline of temperature in the tank, such as after midnight. This leads to a major drawback in the current operating system, particularly when the heat pump produces hot water that is not used by the household. Consequently, this results in wasted energy in the surroundings and compromises user comfort by providing hot water that is not immediately needed.

Hot water consumption patterns in households are influenced by multiple factors, with weather conditions playing a significant role. For example, \cite{Ahmad2020} shows that outdoor temperature affects both the thermal efficiency of heat pumps and user behavior regarding hot water usage. This weather dependency adds another layer of complexity to demand forecasting, as it requires consideration of both short-term weather variations and seasonal patterns. However, while outdoor temperature represents a key factor, it is insufficient for accurate performance \cite{Song2023}. Therefore, it is necessary to incorporate multiple input features in the ML model to effectively estimate heat pump energy consumption patterns.

Building on this, our paper proposes an adaptive approach to managing hot water production in heat pumps, based on household-specific consumption patterns. Unlike the traditional threshold-based method, this approach customizes hot water production to reflect the consumption behavior of each individual household, enhancing both efficiency and user comfort.

\section{The proposed approach for ML-based steering of household hot water production}\label{sec:approach}
The main objective of this study is to design and implement an ML-based steering approach that enhances the efficiency and responsiveness of a household HP system by forecasting the future household demand for hot water. The proposed approach replaces the conventional threshold-based method that steers the hot water production in an HP. This is achieved by predicting $t_\mathrm{mid}$ for a specific time window using ML, then applying an anomaly detection method to build the household demand calendar and steer the production of hot water in an HP. The integration of these methods, predictive ML for accurate temperature prediction, and anomaly detection to identify significant deviations from normal temperature patterns, enables adaptive management of the hot water production in a heat pump.

\begin{figure}
    \centering
    \includegraphics[width=1\linewidth]{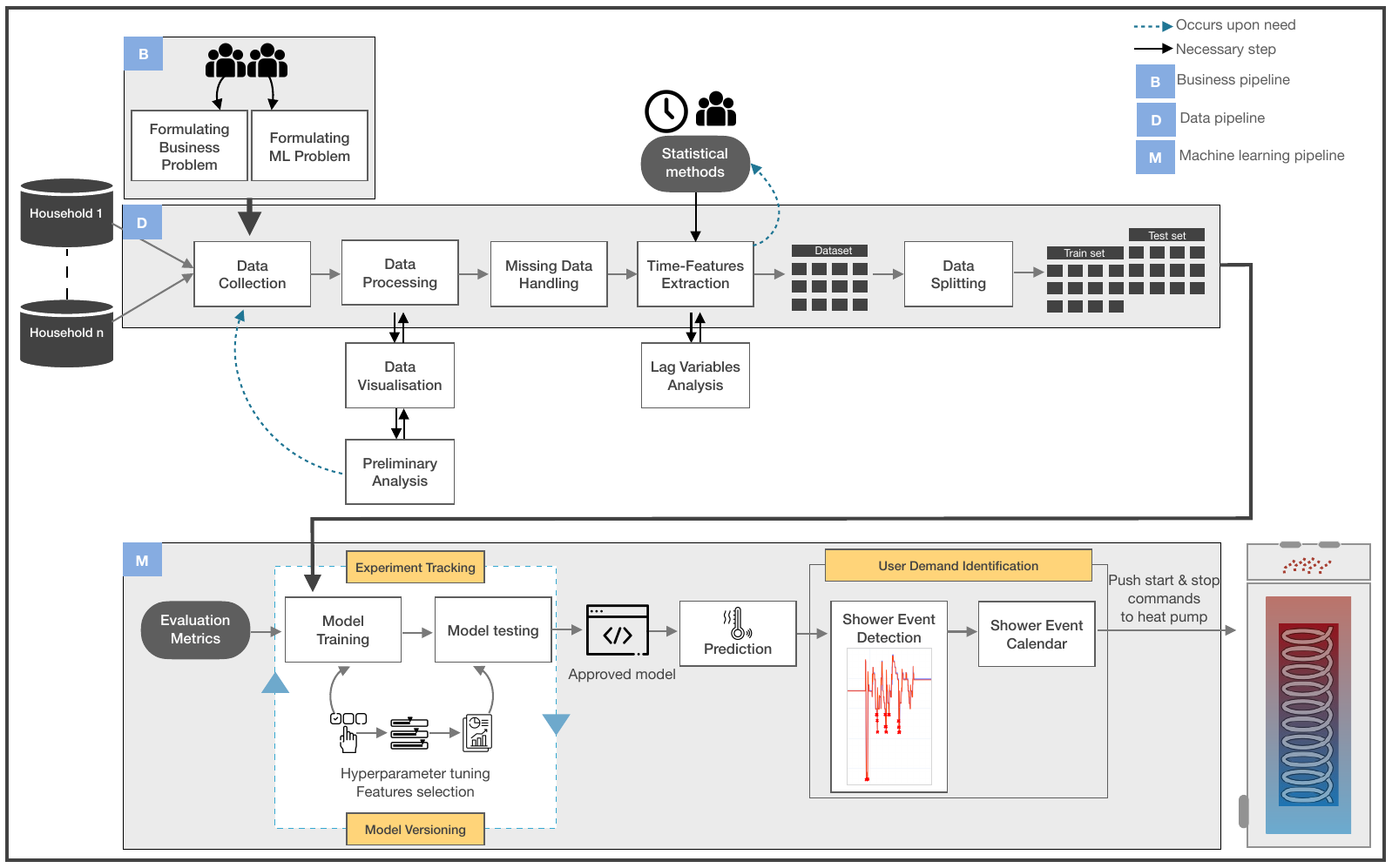}
    \caption{Framework for forecasting and adapting the production of hot water production to the household demand.}
    \label{fig:framework}
\end{figure}

The proposed approach is detailed in Figure \ref{fig:framework}, which illustrates the overall architecture and provides detailed guidance steps. 
Our proposed approach encompasses several steps within three main pipelines: business, data, and ML. The process begins with formulating and translating the business objectives into specific ML tasks. In the data pipeline, data from each household are collected individually, including sensor readings, followed by initial data visualizations and analysis to identify patterns or trends. The raw data for each household are then processed to address inconsistencies and missing values using methods such as forward filling. Once the data are processed, time-based features are extracted to capture temporal trends, in addition to statistical feature analysis. 

The subsequent steps include model training, hyperparameter tuning, and selection of evaluation metrics. Various statistical methods are applied at this stage to identify the most influential features and optimize the model parameters. Model training is used to learn the individual household consumption pattern during the previous year, while model testing evaluates the effectiveness of predictions on the test set. From a deployment perspective, where multiple models are trained and evaluated, experiment tracking and model versioning are crucial for maintaining a record of different model iterations and their performance. Once the best model with the lowest prediction error is identified, the approved model is pushed to the repository and used to predict the values of $t_\mathrm{mid}$ for the selected time window. The model's accuracy and precision are assessed using the chosen evaluation metrics, and the final model that meets performance criteria is approved for deployment. 

In the last stage, iForest, as an anomaly detection method, is used to predict expected shower events. Shower events are detected by identifying significant drops in mid-temperature $t_\mathrm{mid}$ and used to construct a shower event calendar that models the household behavior over time. As the shower event calendar is constructed, start and stop commands are pushed to the HP to steer the hot water production function.
In principle, the demand for thermal energy is dependent on the habits of each household's inhabitants. Their habits can be represented in periodic, time-related information such as time of day, day of week, and holidays. In our case, only time-of-day and day-of-week information is used. The household data characteristics and the randomness factor of a household behavior are motivated by using probabilistic metrics to report the household calendar.

\section{Methodology}\label{sec:methodology}

This section describes the handling of datasets, including preprocessing and extraction of relevant time features. In addition, we present models and methods used at various steps of the proposed approach, along with the rationale behind selecting them for predicting hot water consumption in a household and detecting shower events. This section presents our integrated approach for optimizing heat pump operations through machine learning and anomaly detection. We first describe our model selection rationale and design considerations, followed by a detailed explanation of each model with supporting equations and their justifications. The methodology proceeds sequentially from feature-rich models (LightGBM) through sequential processing models (LSTM and its variants) to anomaly detection methods (iForest), creating a comprehensive framework for accurate hot water demand forecasting and shower event detection. The chosen methods in our approach serve two key purposes. LightGBM is highly efficient for handling large, high-dimensional datasets and is well-suited for IoT systems. Meanwhile, LSTM excels in short-term prediction tasks for time series data. This section also covers dataset handling, including preprocessing techniques and extraction of relevant time features that feed into our sequential modeling pipeline.

\subsection{Model Selection and Design}

The selection of LightGBM and LSTM variants for comparative analysis is informed by several factors supported in recent literature. LSTM models have shown exceptional accuracy in energy forecasting tasks \cite{Alselwi2024} and, specifically, have outperformed other algorithms in short-term load forecasting for individual residential households \cite{Kong2017}. LightGBM is selected for its proven efficiency in IoT systems and its superior performance under time and resource constraints \cite{Yang2021, Mishra2023}. The addition of attLSTM provides insights into whether attention mechanisms can further improve the prediction accuracy for 
household consumption patterns. While other methods exist, these three models represent different approaches (gradient boosting and deep learning), yet have established performance in energy-related time series applications.

Our model selection process follows a systematic approach that considers both computational efficiency and prediction accuracy requirements for residential heat pump systems. The models are presented in order of increasing complexity in terms of temporal data handling, from ensemble methods to recurrent neural architectures with attention mechanisms.

\subsubsection{Justification for Model Selection}
While Transformer-based architectures, such as Time-Series Transformer and Temporal Convolutional Networks (TCN), have shown promising results in various time series forecasting tasks, our experimental design deliberately focused on LSTM variants and LightGBM for several important reasons.

First, the computational resource requirements of residential heat pump systems are a critical consideration. Transformer models, while powerful, require significant computational resources both for training and inference, making them less suitable for edge deployment scenarios common in residential settings. Our selected models offer a better balance between performance and computational efficiency, which is essential for practical implementation in household heat pump controllers.

Second, our preliminary experiments indicated that for the specific task of hot water demand forecasting with our dataset characteristics (moderate time series length with clear daily and weekly patterns), the additional complexity of Transformer architectures did not yield proportional improvements in predictive accuracy. The recurrent nature of LSTM and its variants proved particularly effective at capturing the temporal dependencies in household hot water consumption patterns, while LightGBM excelled at handling the complex non-linear relationships between various input features.

Third, a key objective of our work was to develop models that could be easily interpreted and adapted for specific household patterns. The relative interpretability of our selected models, particularly LightGBM, offers practical advantages for system engineers and users compared to the more complex attention mechanisms in Transformer models.

For completeness, we conducted limited experiments with a Time-Series Transformer model on a subset of our data. The model achieved comparable results to our LSTM variants (mean RMSE of 0.92 compared to LSTM's 0.89) but required approximately 3.5 times longer training time and 2.7 times more memory during inference. These results further validated our model selection choices for the specific task of household hot water demand forecasting.

Future work could explore the application of optimized, lightweight Transformer-based architectures as residential computing capabilities continue to improve and as larger-scale datasets become available.

\subsubsection{LightGBM}

The lightGBM is a model that not only inherits the advantages of a gradient boosting tree but is faster and more lightweight \cite{Zhao2023}. Gradient boosting is an ensemble technique that builds models sequentially, where each new model is trained to correct the errors of its predecessors. Mathematically, if we denote the prediction at step $m$ as $F_m(x)$, the gradient boosting algorithm builds the next model using Equation \ref{Eq:lightGBM} \cite{Zhao2023}

\begin{equation}
    F_{m+1}(x) = F_m(x) + \eta \cdot h_m(x) \label{Eq:lightGBM}
\end{equation}

where $h_m$ is a weak learner trained on the negative gradient of the loss function with respect to the current predictions, and $\eta$ is the learning rate. The improvement comes from selecting the node with the highest gain in its decision tree leaf growth strategy \cite{che2021}. Thanks to gradient-based one-side sampling (GOSS) and exclusive feature bundling (EFB), lightGBM is highly efficient for large and high-dimensional datasets \cite{Yang2021}. GOSS retains instances with large gradients and randomly samples instances with small gradients to maintain accuracy while reducing computational cost. EFB bundles mutually exclusive features to reduce dimensionality without losing information, making it particularly efficient for sparse features common in IoT systems. LightGBM has proven suitable for IoT systems due to its efficiency under time and resource constraints \cite{Yang2021}, making it a suitable choice for applications such as HP systems. Therefore, the implementation of lightGBM in our paper is motivated by the ensemble-based methods used in IoT systems, where it has demonstrated superior performance compared to other ML models \cite{Mishra2023}.


\subsubsection{LSTM}
The main goal in time series forecasting is to identify non-trivial patterns in historical data to predict useful future information about a certain time-correlated phenomenon \cite{Parmezan2019}. Time series data can exhibit non-linear patterns that require using non-linear models for forecasting tasks \cite{Abbasimehr2020}. LSTM networks address the vanishing gradient problem in traditional recurrent neural networks through a memory cell architecture with three main gates: input, forget, and output gates. The LSTM cell state $C_t$ and hidden state $h_t$ at time $t$ are updated based on Equations \ref{Eq2}-\ref{Eq7} \cite{Hochreiter1997}.

\begin{equation}
i_t = \sigma(W_i \cdot [h_{t-1}, x_t] + b_i)\label{Eq2}
\end{equation}

\begin{equation}
f_t = \sigma(W_f \cdot [h_{t-1}, x_t] + b_f)\label{Eq3}
\end{equation}

\begin{equation}
o_t = \sigma(W_o \cdot [h_{t-1}, x_t] + b_o)\label{Eq4}
\end{equation}

\begin{equation}
\tilde{C}_t = \tanh(W_c \cdot [h_{t-1}, x_t] + b_c)\label{Eq5}
\end{equation}

\begin{equation}
C_t = f_t * C_{t-1} + i_t * \tilde{C}_t \label{Eq6}
\end{equation}

\begin{equation}
h_t = o_t * \tanh(C_t) \label{Eq7}
\end{equation}

where $x_t$ is the input at time $t$, $\sigma$ is the sigmoid function, $*$ denotes element-wise multiplication, and $W$ and $b$ are the weight matrices and bias vectors, respectively. The nonlinearity and adaptability of an NN and its variants motivated their wide use in forecasting energy-related data \cite{Wei2018}. Of these variants, LSTM is widely used given its ability to minimize the likelihood of the vanishing gradient problem and recall values from previous stages compared to classic recurrent neural networks (RNNs) \cite{Hochreiter1997, Farzad2019}.

Two main components were added to the simple RNN architecture: cell states, which store past information, and gates, which control the information flow \cite{alselwi2023}. Additionally, LSTM includes hidden states to represent short-term memory. Given that LSTM networks are efficient for modeling time-variant systems \cite{Lindemann2021}, they are widely used to handle energy-related problems. They offer high performance in various applications, such as forecasting electricity load and energy prices \cite{Alselwi2024}. Therefore, LSTM is one of the models that is selected to forecast household demand for hot water consumption.

\subsubsection{Bi-directional LSTM}
The complex structure of LSTM networks makes them ideal for time series problems but prone to gradient explosion or vanishing issues \cite{Cui2024}. Extensions like Bi-directional LSTM (BiLSTM) address these challenges by incorporating both past and future information during training, unlike standard LSTM which only uses past information. However, BiLSTM networks require longer training times \cite{Shahmohammadi2021}. The computational complexity can be addressed by using an attention mechanism that helps the model to consider the most important information from the input data \cite{Zhou2016attention}. 

The self-attention mechanism calculates the weighted importance of different input elements using Equation \ref{eq:attention} \cite{Zhou2016attention}

\begin{equation}
    \textcolor{blue}{\text{Attention}(Q, K, V) = \text{softmax} \left( \frac{QK^T}{\sqrt{d_k}} \right) V}
    \label{eq:attention}
\end{equation}

where \( Q, K, V \) are the query, key, and value matrices, and \( d_k \) is a scaling factor that prevents large gradient values. This mechanism improves model performance by dynamically adjusting the weight of different time steps in the sequence. 

The attention mechanism improves BiLSTM by determining the weight of each data position, thus producing the final refined outputs \cite{Cui2024}.

 In the literature, many studies address short-term forecasting tasks with robust vanilla LSTM models; however, few consider the promising potential of BiLSTM \cite{Feng2019}, particularly in forecasting thermal energy demand. This paper tests a BiLSTM-based forecasting structure combined with a self-attention mechanism. AttLSTM is expected to improve the accuracy of the model, as BiLSTM can offer superior performance for short-term prediction tasks \cite{Abduljabbar2021}. With the purpose of generating a relatively accurate calendar of hot water demand for every household, providing a high forecast performance of $t_\mathrm{mid}$ is crucial before running the next task of shower detection. The BiLSTM model with the attention mechanism (attLSTM) was chosen to enhance the model’s ability to capture long-term dependencies by processing the input data in both forward and backward directions. The attention mechanism further improves performance by assigning higher importance to more relevant time steps, allowing the model to focus on critical information. However, a limitation of this approach is its increased computational complexity compared to simpler models such as LSTM and LightGBM. This additional complexity may result in longer training times and higher memory consumption, which could be a concern for real-time applications.
 
 \subsubsection{IForest for shower Event Detection}
 
In this context, a shower event is defined as a significant drop in temperature over a short period of time, indicating substantial use of hot water and a deviation from typical usage patterns. Our definition agrees with the characteristics of anomalous data described by Sahand \emph{et al.} that such data are few and different \cite{Sahand2021}. Additionally, an anomaly is defined as the collection of anomalous data instances with respect to the normal data behavior \cite{Chandola2009}. In another definition by Boniol \emph{et al.}, anomalies can correspond to the actual behavior of interest in a measured system \cite{Boniol2023}. Inspired by these definitions, a shower event is a large drop in $t_\mathrm{mid}$ recorded by the sensor placed in the middle of the tap water tank within an HP, referred to as the mid-sensor. These observed drops are termed "shower events." Therefore, an anomaly detection method is to identify shower events in the recorded historical data of households.

Anomaly detection methods are widely used for detecting faults and other applications in HVAC systems. However, the main challenge is choosing the threshold that separates the anomaly event from other events \cite{Mirnaghi2020}. The time dependence and dynamic nature of the time series data make detecting anomaly behavior based on threshold-based decisions a real challenge. Especially in the context of predicting household behavior, given that every household has a unique consumption pattern depending on the habits and size of the family. To overcome this challenge, iForest which operates on a distinct principle compared to the statistical and clustering approaches is selected to detect shower events \cite{Chabchoub2022}. Furthermore, other existing anomaly detection methods such as Density-Based Spatial Clustering of Applications with Noise (DBSCAN) suffers from high complexity in high-dimensional and large datasets, as the main memory is needed to load the whole dataset \cite{Viswanath2006}. Similarly, one-class SVM requires high computational capacity in large datasets and is considered sensitive to the the kernel function parameter tuning \cite{Kang2019}. As for autoencoders (AEs), which are being widely used in anomaly detection, AEs can suffer from lack of robustness as they are sensitive to noisy data, varied input and outliers \cite{berahmand2024}. Unlike the above mentioned anomaly detection methods, iForest does not depend on selecting thresholds for samples and is well-suited to large datasets . 

Instead, iForest assigns an anomaly score to each instance based on the expected path length within isolation trees as in Equation \ref{Eq:9} \cite{Liu2008}

\begin{equation}
    \textcolor{blue}{s(x, n) = 2^{-\frac{E(h(x))}{c(n)}}}\label{Eq:9}
\end{equation}

where \( E(h(x)) \) is the expected path length of instance \( x \), and \( c(n) \) is the average path length of unsuccessful searches in a binary tree. A lower expected path length corresponds to a higher anomaly score, enabling the identification of shower events.

An iForest algorithm uses an ensemble of decision trees to isolate anomalies based on a preset contamination ratio of the training set \cite{Vitorino2022}. Another advantage of iForest is its low memory requirements and excellent scalability performance that fits large datasets \cite{Liu2008, Chabchoub2022}. Given the large number of households and their varying consumption patterns, using IForest ensures scalable performance across diverse datasets. Prioritizing the lightweight iForest over other methods enhances resource efficiency by minimizing the computational load on the cloud hosting the ML system. Additionally, this choice improves responsiveness by ensuring low latency in communication between the ML system and the HP, particularly after periodic model re-training. The contamination rate chosen for all the use cases in this paper is 0.05. Despite the theoretical suitability of the approach, the results showed an over-detection of shower events. Therefore, post-processing of the detected shower events is implemented to avoid the detection of consecutive shower events within a window of 30 minutes.

Having established the theoretical framework and models, we now describe the specific household datasets used to implement and validate our approach.

\subsection{\textcolor{blue}{Household dataset}}\label{sec:dataanalysis}
The data preprocessing methods can significantly affect the performance of the ML model \cite{Khalaji2023}. In this paper, the selected data preprocessing techniques are motivated by the nature of the acquired time series data, where the values recorded by the sensors are updated at specific intervals, resulting in gaps. Therefore, a comprehensive investigation is conducted to prepare the data for ML modeling by detecting and correcting incomplete and inaccurate data values. Section \ref{sec:dataanalysis} describes the dataset and the data handling techniques used, in addition to the selection and generation of time-step input predictors.

\subsubsection{Collection and preprocessing of household data}\label{sec:Dataset}
This section outlines the characteristics of the data used to test and validate our proposed methods. The datasets correspond to six real HP installations in residential units in Sweden. They vary in type, either GSHP or ASHP and in technical characteristics, such as compressor's effect size. 

The sensor installation and data collection system consists of two temperature sensors installed in the water tank ($t_{\mathrm{top}}$ and $t_{\mathrm{mid}}$) as illustrated in Figure~1. Data recording follows a specific frequency where values are stored every minute, but only in case of change. The collected data includes comprehensive information related to the operation of tap water, including temperature readings from both sensors, compressor power, the in and out condenser temperature, variables related to the starting and stopping temperature thresholds, and type of HP and other specific characteristics applicable to the manufacturer and operating system.

The criteria for selecting the use case data are meticulously curated to ensure a comprehensive and consistent basis for analysis. Specifically, each dataset should include at least one year of training data from all involved households. Additionally, the selected households represent various types of HPs, all running on the same operating software but exhibiting different consumption behaviors. Detailed information on the dataset is shown in Table \ref{tab:dataset_info}. The table presents the start and end dates for the training and test dataset, divided by 85\% and 15\%, respectively. The split is chosen so that the training and validation dataset contains the entire yearly data for each customer while keeping a reasonable amount to be used for testing.

\begin{table}
\centering
\caption{Information of households dataset}
\begin{tabular}{lcccccc}
\toprule
\textbf{Household}& 
\multicolumn{3}{c}{\textbf{Training and validation dataset}} & 
\multicolumn{3}{c}{\textbf{Test dataset}} \\
\cmidrule(lr){2-4} \cmidrule(lr){5-7}
& Start date & End date & Size & Start date & End date & Size \\
\midrule
\textbf{Household 1} & 14.10.2022 & 27.09.2023 & 472945 & 27.09.2023 & 24.11.2023 & 83461 \\
\textbf{Household 2} & 27.10.2022 & 26.09.2023 & 481219 & 26.09.2023 & 24.11.2023 & 84921 \\
\textbf{Household 3} & 24.10.2022 & 06.09.2023 & 456071 & 06.09.2023 & 24.11.2023 & 114018 \\
\textbf{Household 4} & 03.11.2022 & 27.09.2023 & 472588 & 27.09.2023 & 24.11.2023 & 83399 \\
\textbf{Household 5} & 09.12.2022 & 03.10.2023 & 427991 & 03.10.2023 & 24.11.2023 & 75528 \\
\textbf{Household 6} & 29.10.2022 & 26.09.2023 & 478737 & 26.09.2023 & 24.11.2023 & 84484 \\
\bottomrule
\end{tabular}
\label{tab:dataset_info}
\end{table}

The six household data are of dimensions \emph{n} x \emph{m}, where \emph{n} denotes the timestamp records and \emph{m} represents the sensors that are expected to influence hot water production (\emph{m} = sensors and other predictors). In general, the data collected contains information related to the operation of tap water including the compressor power, the different heat sensors installed in the water tank, the in and out condenser temperature, and other variables related to the starting and stopping temperature thresholds. Other variables collected include the type of HP and other specific characteristics applicable to the manufacturer and operating system. The collected data are historical, covering recordings from 2022/2023 with a granularity of one second. However, for easier analysis and visualization, the data is down-sampled per minute.

To perform preliminary analysis, a combination of statistical tools is applied to explore the empirical data distribution and assess data quality with respect to shape, skewness, and outliers. The data quality assessment revealed two key challenges that could affect the performance of the ML models, missing data in the time series and outliers. Missing data primarily occurred due to the data collection approach, where sensor values are stored every minute, but only in case of change. To address this challenge systematically, the missing timestamp measurements in the data were rectified using a forward-fill approach. This approach propagates the last valid observation forward to fill in missing values \cite{Ribeiro2022}. In addition, data were resampled to 1-minute granularity for consistency, and maintaining temporal continuity. For outlier detection and handling, box plots were used to identify temperature values outside the expected ranges, where data outage periods were discarded. Additionally, violin plot was used to illustrate the density shape of the numerical data \cite{Blumenschein2020}. The forward fill method was chosen for handling missing values due to its simplicity and effectiveness in time-series data where the most recent valid observation is likely to be similar to the missing data point. While alternative methods such as interpolation and statistical imputation exist, forward fill is particularly suitable for time-series forecasting where continuity of past measurements is essential. The accuracy of this method was validated through visual inspection and comparison against known valid sequences to ensure continuity in temperature readings.

\begin{figure}
    \centering
    \includegraphics[width=0.5\linewidth]{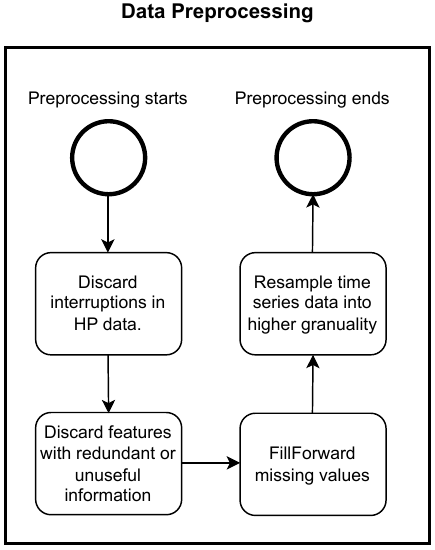}
    \caption{Overview of the household data preprocessing steps.}
    \label{fig:preprocessing}
\end{figure}

These tools enabled us to distinguish between genuine usage patterns and technical faults in sensor readings. In the next step, variables that are constant or exhibit low variation were discarded , as they are not useful for the learning task. Therefore, 11 out of 23 independent variables that exhibit a constant nature were removed from the dataset. This procedure is similarly applied to all six datasets. A summary of the main preprocessing steps is illustrated in shown in Figure \ref{fig:preprocessing}. 

\begin{figure}
    \centering
    \subfigure[Household 1]{
        \resizebox*{0.47\textwidth}{!}{\includegraphics{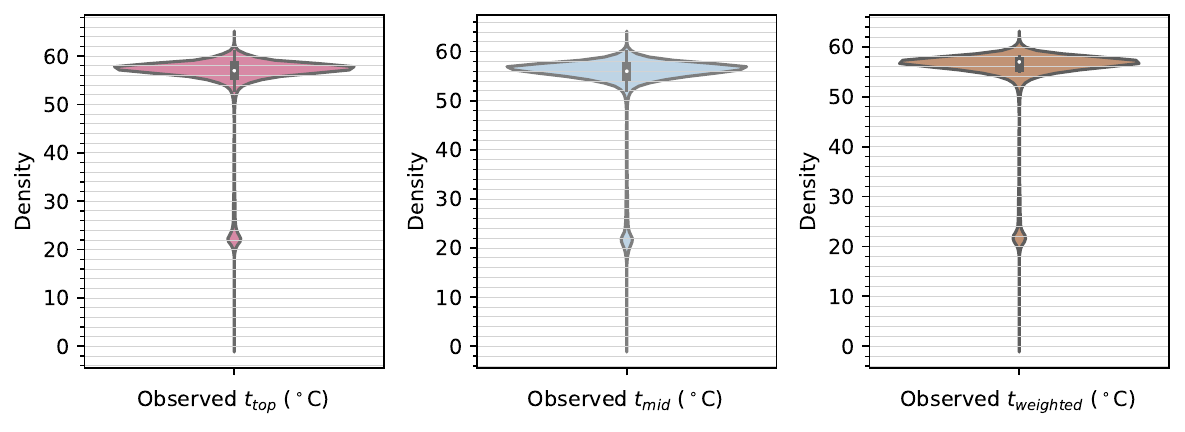}}}\hspace{8pt}
    \subfigure[Household 2]{
        \resizebox*{0.47\textwidth}{!}{\includegraphics{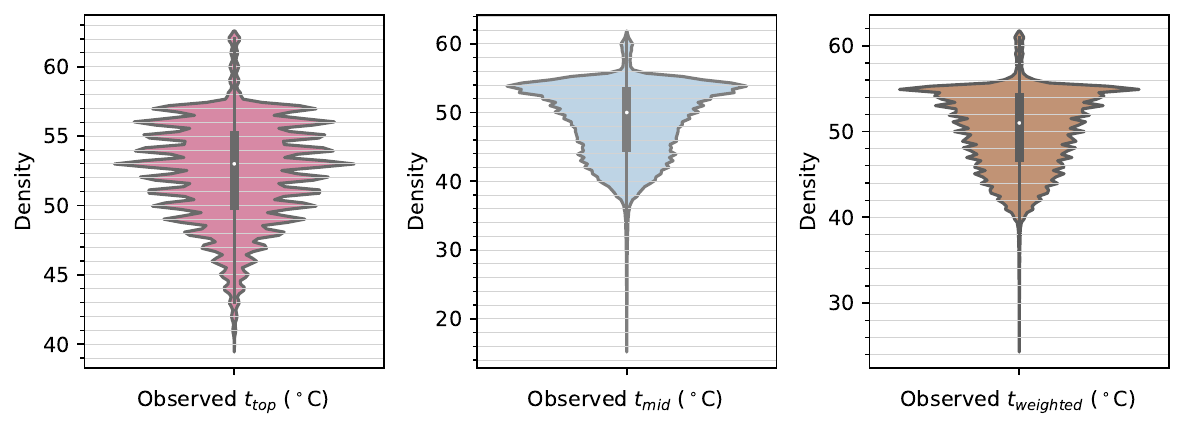}}}
    \vspace{10pt}
    \subfigure[Household 3]{
        \resizebox*{0.47\textwidth}{!}{\includegraphics{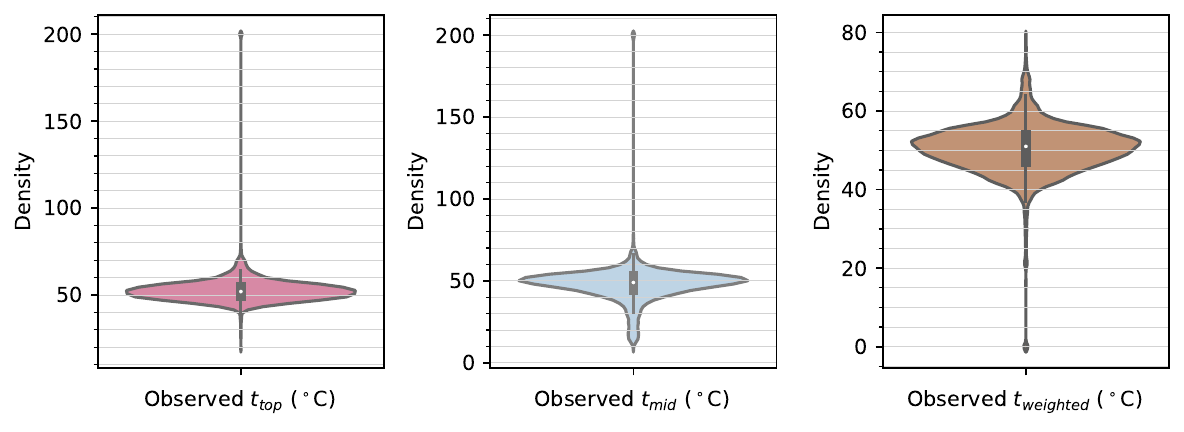}}}\hspace{8pt}
    \subfigure[Household 4]{
        \resizebox*{0.47\textwidth}{!}{\includegraphics{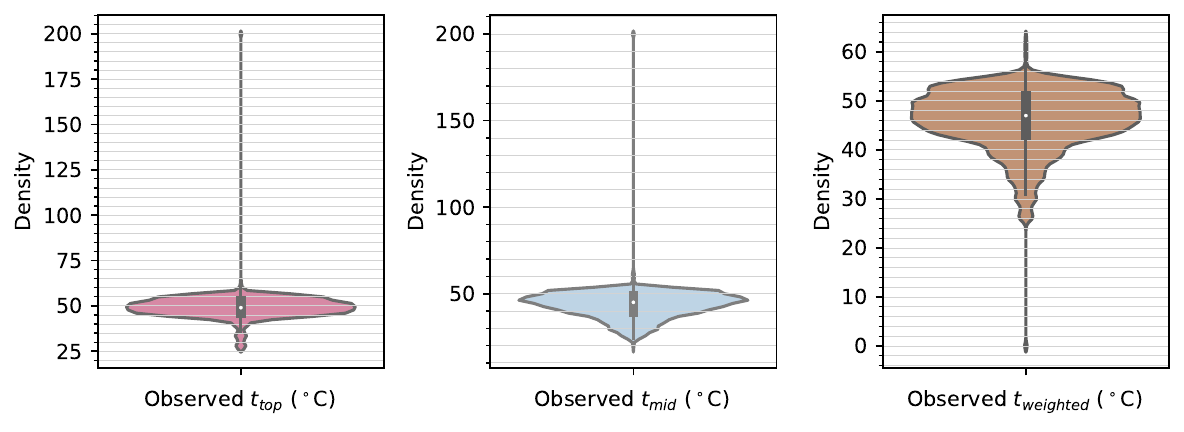}}}
    \vspace{10pt}
    \subfigure[Household 5]{
        \resizebox*{0.47\textwidth}{!}{\includegraphics{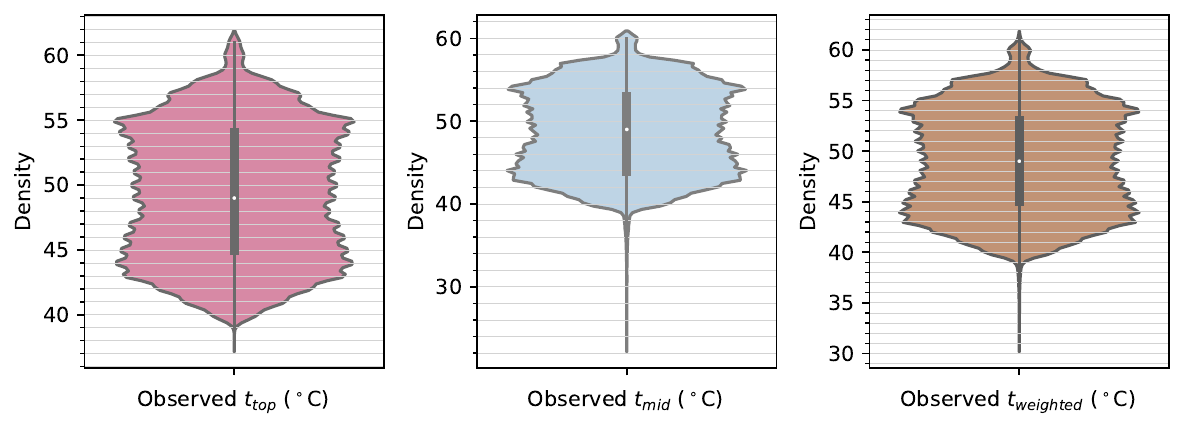}}}\hspace{8pt}
    \subfigure[Household 6]{
        \resizebox*{0.47\textwidth}{!}{\includegraphics{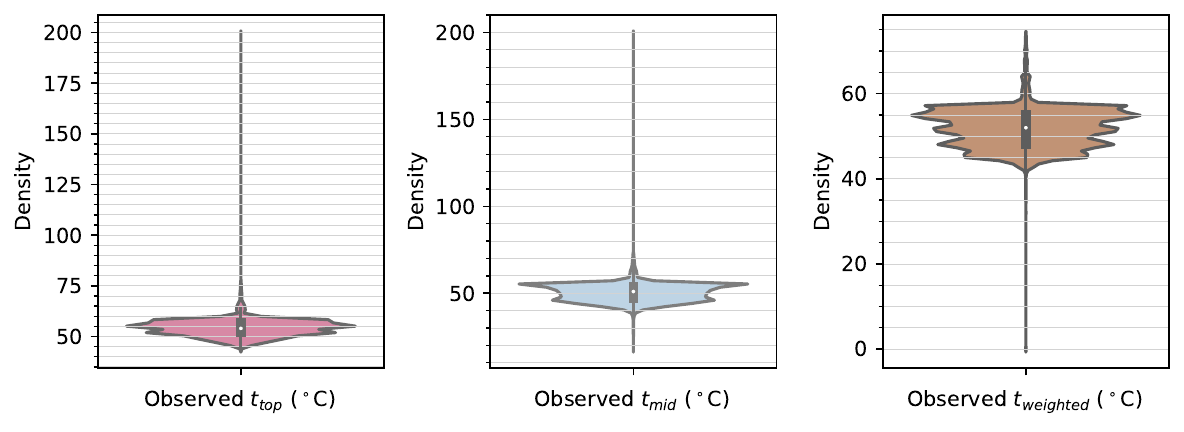}}}
    \caption{Distribution of measurements of sensors per household}
    \label{fig:hpviolin}
\end{figure}

The complexity of the data is mainly attributed to the size of the observations and the variables \cite{Wichitaksorn2023}. Advancements in computational technologies have made it easier to analyze big data. However, the large number of variables still introduces complexity to data analysis \cite{Wichitaksorn2023}. Hence, there is an advantage to reducing the number of variables through the feature selection process. After filtering out variables with low variation and categorical variables, a large space of independent variables remains, including collected and retrieved time-related variables. To further narrow the feature space, ordinary least squares (OLS) regression is performed to understand the relationship between the remaining variables and the outcome of interest, $t_\mathrm{mid}$. The output p-value associated with the independent variables shows that most of the selected variables are statistically significant but with a negligible effect size except for $t_\mathrm{top}$. Therefore, $t_\mathrm{top}$ is selected as one of the influential input variables. In Figure \ref{fig:hpviolin}, the distribution of measurements of the observed $t_\mathrm{top}$, $t_\mathrm{mid}$, and $t_\mathrm{weighted}$ per household is shown, where $t_\mathrm{weighted}$ represents the average temperature of $t_\mathrm{top}$ and $t_\mathrm{mid}$.

\subsubsection{Generation of time-steps input predictors}\label{sec:featuregenerate}
After cleaning the data and retaining a single predictor, the next step is to check the significance of time-generated predictors (lag variables). The predictors generated from the available timestamp can significantly influence the model's performance. The household data are collected with a timestamp in dd-mm-yyyy hh: mm: ss format. Six features, including day of the year, month, day of the week, hour of the day, minute of the hour, and holiday, are deduced and tested in the first stage. As a result of applying the OLS regression method, particularly observing the p-value for every new feature, most time-related variables showed no significance except for week and month. In this case study, only the week information is retained to avoid redundancy and maintain a variable with lower granularity, representing the temporal context.

The primary purpose of analyzing time series data is to uncover patterns from past observations \cite{Liu2020}, and the lag variables contribute to the discovery of these patterns. Therefore, in the second feature analysis stage, the problem is framed as a multi-step forecasting problem that utilizes pre-selected and influential lag variables as input. The significance of lag variables on the $t_\mathrm{mid}$ is investigated, particularly ten-minute intervals up to two hours. Using partial OLS regression, the lag temperature values recorded at specific intervals (90, 30, 20, and 10 minutes prior) for both $t_\mathrm{mid}$ and $t_\mathrm{top}$ are significant and provide a temporal context for the model prediction performance. The lag variables at 90, 30, 20, and 10 minutes are found to be statistically significant for all six household data. As a result, lags at times 10, 20, 30, and 90 minutes are chosen to be included as input for the ML model, as shown in Table \ref{tab:features}. Given the significance of $t_\mathrm{top}$, the current $t_\mathrm{top}$ is also included to provide the model with the latest state of the water tank temperature profile. As for the week predictor, it is included as an integer to account for seasonal variations that may influence the consumption of hot water in a household. The outcome predictors collectively help the ML model to accurately forecast $t_\mathrm{mid}$ and capture the temporal consumption behavior.

\begin{table*}
\centering
\caption{ML selected input feature subset including lag variables \textcolor{blue}{at a previous time}.}
\label{tab:features}
\begin{tabular}{lll}
\toprule
\textbf{Feature} & \textbf{Type} & \textbf{Description}\\
\midrule
\textbf{t$_\mathrm{mid}$(90)} & float &lag temperature value at the mid of the water tank ($^\circ$C) at minute 90\\
\textbf{t$_\mathrm{mid}$(30)} & float &lag temperature value at the mid of the water tank ($^\circ$C) at minute 30\\
\textbf{t$_\mathrm{mid}$(20)} & float &lag temperature value at the mid of the water tank ($^\circ$C) at minute 20\\
\textbf{t$_\mathrm{mid}$(10)} & float &lag temperature value at the mid of the water tank ($^\circ$C) at minute 10\\
\textbf{t$_\mathrm{top}$(90)} & float &lag temperature value at the top of the water tank ($^\circ$C) at minute 90\\
\textbf{t$_\mathrm{top}$(30)} & float &lag temperature value at the top of the water tank ($^\circ$C) at minute 30\\
\textbf{t$_\mathrm{top}$(20)} & float &lag temperature value at the top of the water tank ($^\circ$C) at minute 20\\
\textbf{t$_\mathrm{top}$(10)} & float &lag temperature value at the top of the water tank ($^\circ$C) at minute 10\\
\textbf{t$_\mathrm{top}$} & float &lag temperature value at the top of the water tank ($^\circ$C) at time \emph{t}\\
\textbf{Week} & Integer &Number of week in the year\\
\bottomrule
\end{tabular}
\end{table*}

The selected features for this study, including historical temperatures, current temperature, and time-based features, were chosen based on their practical relevance for hot water consumption forecasting. While an ablation study could provide further insights into the individual impact of each feature, it was not conducted as this study primarily focuses on evaluating comprehensive model performance. Future work could explore an ablation study to better understand the contribution of each feature.

\section{Implementation and Empirical Evaluation}\label{sec:results}
To ensure a fair comparison, three models are trained, an LSTM model, lightGBM, and bi-LSTM model, applying them individually to each of the household data. The list of extracted features presented in Table \ref{tab:features} is similarly used for running all three models. The performance of the deployed ML models is evaluated using records from real-installed HPs from 6 households. The preprocessed data used in modeling are divided into 85\% for training and 15\% for testing. Particularly for the LSTM-based methods, the structure of the model is composed of an LSTM layer combined with a dense layer.

\subsection{Experimental setup}
To gain insight into the computational environment for implementing our approach, we include in this paper Table \ref{tab:environment} which presents the specifications for the local machine used for light preprocessing tasks, and the corresponding server instance used for modeling.


\begin{table}
\centering
\caption{Computational specifications}\label{tab:environment}
\begin{tabular}{lll}
\toprule
\textbf{} & \textbf{Local environment} & \textbf{Server} \\
\midrule
\textbf{CPU} & Apple M1 Pro & Intel Core i9-9900X CPU \\
\textbf{GPU} & - & Nvidia GeForce RTX 2080, 11GB GDDR6 \\
\textbf{RAM} & 16GB DDR4 & 64GB DDR4 \\
\bottomrule
\end{tabular}
\end{table}

\subsection{Error metrics}
To compare and evaluate the prediction performance of the three ML models used, mean absolute percentage error (MAPE) and the root mean square error (RMSE) are measured, as they are commonly used in time series forecasting problems \cite{Nti2020}.
\begin{equation}
\text{MAPE} = \frac{1}{n} \sum_{i=1}^{n} \left| \frac{y_i - \hat{y}_i}{y_i} \right| \times 100
\end{equation}
\begin{equation}
\text{RMSE} = \sqrt{\frac{1}{n} \sum_{i=1}^{n} (y_i - \hat{y}_i)^2}
\end{equation}
where 
where \( y_i \) represents the vector of the actual $t_\mathrm{mid}$ values, \( \hat{y}_i \) the predicted values, and \( n \) the total number of predictions.

\subsection{Hyperparameter tuning}

The hyperparameter tuning of the LightGBM model is conducted using sklearn's RandomizedSearchCV library to identify optimal parameters for predicting hot water consumption. The optimization objective is to minimize the RMSE across a three-fold cross-validation, where the final selected model demonstrated the lowest RMSE. Table \ref{tab:hypertuning} summarizes the hyperparameter search space used in the experiment.

The selection of hyperparameters for the LightGBM model was guided by both theoretical considerations and empirical experimentation. The choice of hyperparameter search space was determined based on common values used in time series prediction tasks, while also considering computational efficiency. The selection process was conducted using the sklearn's RandomizedSearchCV library, which allows for a more efficient search over a wide range of hyperparameter combinations compared to GridSearchCV. This approach was chosen to reduce the computational cost while still providing robust hyperparameter optimization.

In regard to LSTM tuning, it is performed using the Keras tuner's RandomSearch library, which allows for efficient exploration of hyperparameter combinations. The search range for LSTM units (50 to 150) was determined based on preliminary experiments, which indicated that values outside this range resulted in overfitting or underfitting. Adam was chosen as the optimizer due to its demonstrated efficiency in training deep learning models for time series forecasting. The hyperparameter selection process aimed to minimize validation loss during training, which directly correlates with better generalization performance. The final model is then trained for 50 epochs using the optimal hyperparameters identified. To ensure full reproducibility, Table \ref{tab:hyperparams} provides a comprehensive overview of the final hyperparameters used for each model, along with the predictors and response variable used during training.



\begin{table}
\caption{Hyperparameter Search Space for LightGBM Tuning}
\label{tab:hypertuning}
\centering
\begin{tabular}{ll}
\toprule
\textbf{Model} & \textbf{Parameter Grid} \\
\midrule
LightGBM & \parbox{6cm}{
\begin{itemize}
\item Boosting\_type: \{gbdt, dart\}
\item Max\_depth: \{5, 10, 30\}
\item Learning\_rate: \{0.001, 0.01, 0.1\}
\item N\_estimators: \{100, 500, 800\}
\item Num\_leaves: \{2, 5, 10, 20\}
\end{itemize}
} 
\bottomrule
\end{tabular}
\end{table}

\begin{table}[h]
    \centering
    \caption{Hyperparameters and Predictors Used in Model Training}
    \label{tab:hyperparams}
    \begin{tabular}{|l|l|l|}
        \hline
        \textbf{Model} & \textbf{Hyperparameters} & \textbf{Values} \\ \hline
        \multirow{5}{*}{\textbf{LSTM}} 
        & LSTM Units & 50 \\ \cline{2-3}
        & Optimizer & Adam \\ \cline{2-3}
        & Loss Function & MAE \\ \cline{2-3}
        & Epochs & 50 \\ \cline{2-3}
        & Batch Size & 72 \\ \hline
        \multirow{5}{*}{\textbf{LightGBM}} 
        & Boosting Type & gbdt, dart \\ \cline{2-3}
        & Max Depth & 5, 10, 30 \\ \cline{2-3}
        & Learning Rate & 0.001, 0.01, 0.1 \\ \cline{2-3}
        & Number of Estimators & 100, 500, 800 \\ \cline{2-3}
        & Number of Leaves & 2, 5, 10, 20 \\ \hline
        \textbf{Predictors} & Features Used & tmid(90), tmid(30), tmid(20), tmid(10), \\ 
        & & ttop(90), ttop(30), ttop(20), ttop(10), ttop, Week \\ \hline
        \textbf{Response (Target)} & Predicted Variable & tmid (future temperature at mid-tank) \\ \hline
    \end{tabular}
\end{table}

\subsection{Predictive performance comparison and evaluation} \label{sec:evaluation}

\begin{figure}
    \centering
    \includegraphics[width=1\linewidth]{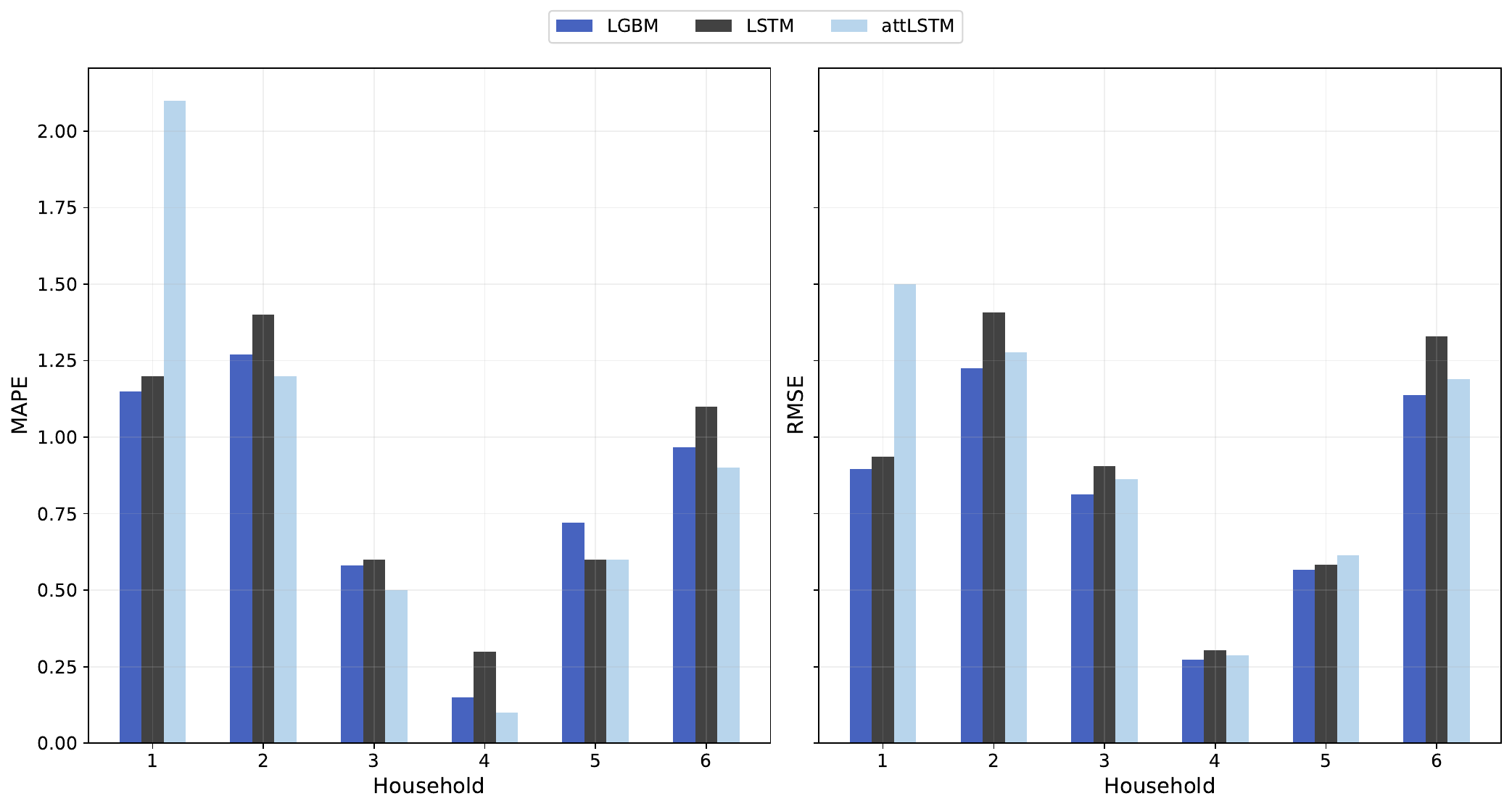}
    \caption{Overall MAPE and RMSE for the six households.}
    \label{fig:error}
\end{figure}

Table \ref{tab:results} shows the prediction error of all households observed for lightGBM, LSTM, and attLSTM. Overall mean MAPE and RMSE performance metrics are calculated for each household across 50 epochs for LSTM and attLSTM. Across all households, the lightGBM model consistently outperforms LSTM variants with lower error rates for the given data. On average, the RMSE for lightGBM is approximately 9.37\% lower compared to LSTM between the different households, with household 6 recording the highest difference between both models. However, we note that lightGBM is hypertuned for every household dataset, confirming that one lightGBM model does not fit different household behaviors. In this context, global models that lack sufficient complexity are likely to be outperformed by local models when using the same set of hyperparameters \cite{Buonanno2022}. Meanwhile, for LSTM, hypertuning results in similar parameters for all six datasets. In terms of households, household 4 consistently shows low RMSE and MAPE values across all models, indicating a clear consumption pattern throughout the period tested, as shown in Figure \ref{fig:error}. Household 5 also performs very well, with low RMSE and MAPE. Household 3 and household 2 have good performance but are not as consistent as Household 4. Households 1 and 6 show higher RMSE and MAPE values than the others. This indicates more randomness in their daily or weekly patterns of hot water consumption. 

\begin{table}
\setlength{\tabcolsep}{3pt}
\centering
\caption{Performance results of all households.}
\label{tab:results}
\begin{tabular}{lccccccccc}
\toprule
\textbf{Household \#} & \multicolumn{3}{c}{\textbf{LGBM}} & \multicolumn{3}{c}{\textbf{LSTM}} & \multicolumn{3}{c}{\textbf{attLSTM}} \\
\cmidrule(r){2-4} \cmidrule(lr){5-7} \cmidrule(lr){8-10}
& \textbf{R²} & \textbf{RMSE} & \textbf{MAPE} & \textbf{R²} & \textbf{RMSE} & \textbf{MAPE} & \textbf{R²} & \textbf{RMSE} & \textbf{MAPE} \\
\midrule
\textbf{Household 1} &0.748  &0.895 &0.011 &0.725& 0.936& 0.012&0.286  & 1.50& 0.021 \\
\textbf{Household 2} & 0.941 & 1.225 & 0.012&0.922 & 1.408 &0.014 &  0.936& 1.277&0.012 \\
\textbf{Household 3} &0.948  &0.814  &  0.005&0.936 & 0.906 & 0.006& 0.942 &0.862&0.005\\
\textbf{Household 4} &0.956  &0.273 & 0.001&0.946  & 0.304 & 0.003& 0.952 & 0.287&0.001 \\
\textbf{Household 5} & 0.983 & 0.567 & 0.007&0.982 & 0.583 & 0.006 &0.98& 0.613& 0.006 \\
\textbf{Household 6} &0.936  &1.138  & 0.009&0.913 & 1.329 &0.011 & 0.93 &1.189&0.009  \\
\bottomrule
\end{tabular}
\end{table}

\begin{figure}
    \centering
    \subfigure[Household 4]{
        \resizebox*{6.8cm}{!}{\includegraphics{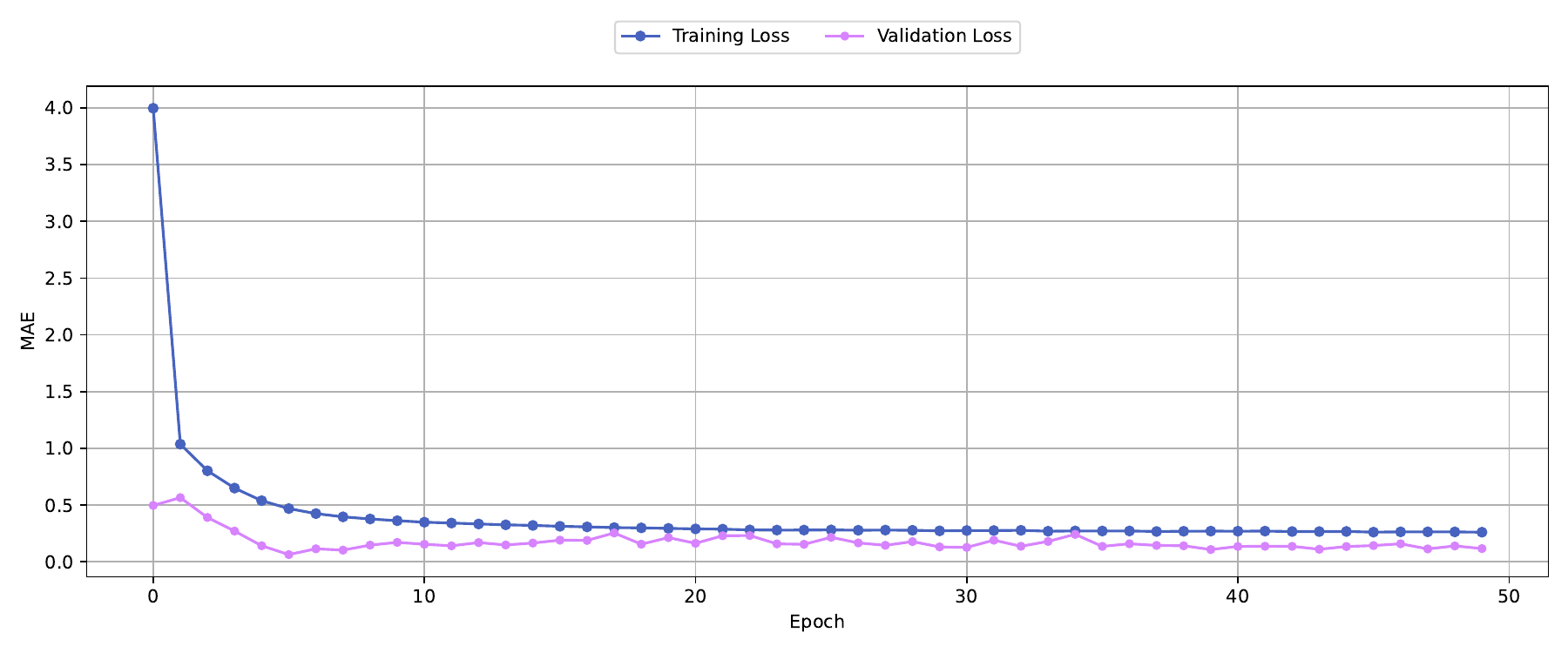}}}\hspace{5pt}
    \subfigure[Household 5]{
        \resizebox*{6.8cm}{!}{\includegraphics{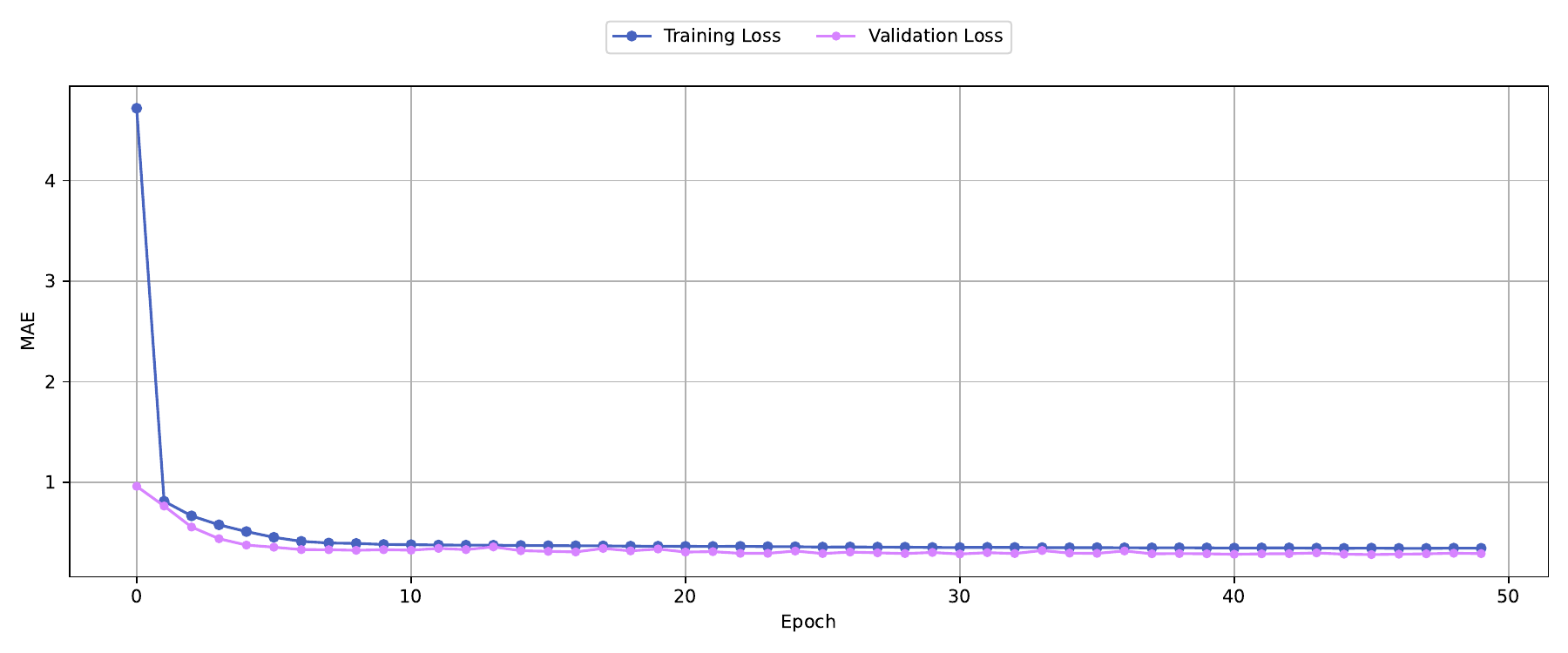}}}
    \caption{LSTM learning curves for household 4 and 5}
    \label{fig:LSTMlosscurves}
\end{figure}

\begin{figure}
    \centering
    \subfigure[Household 4]{
        \resizebox*{6.8cm}{!}{\includegraphics{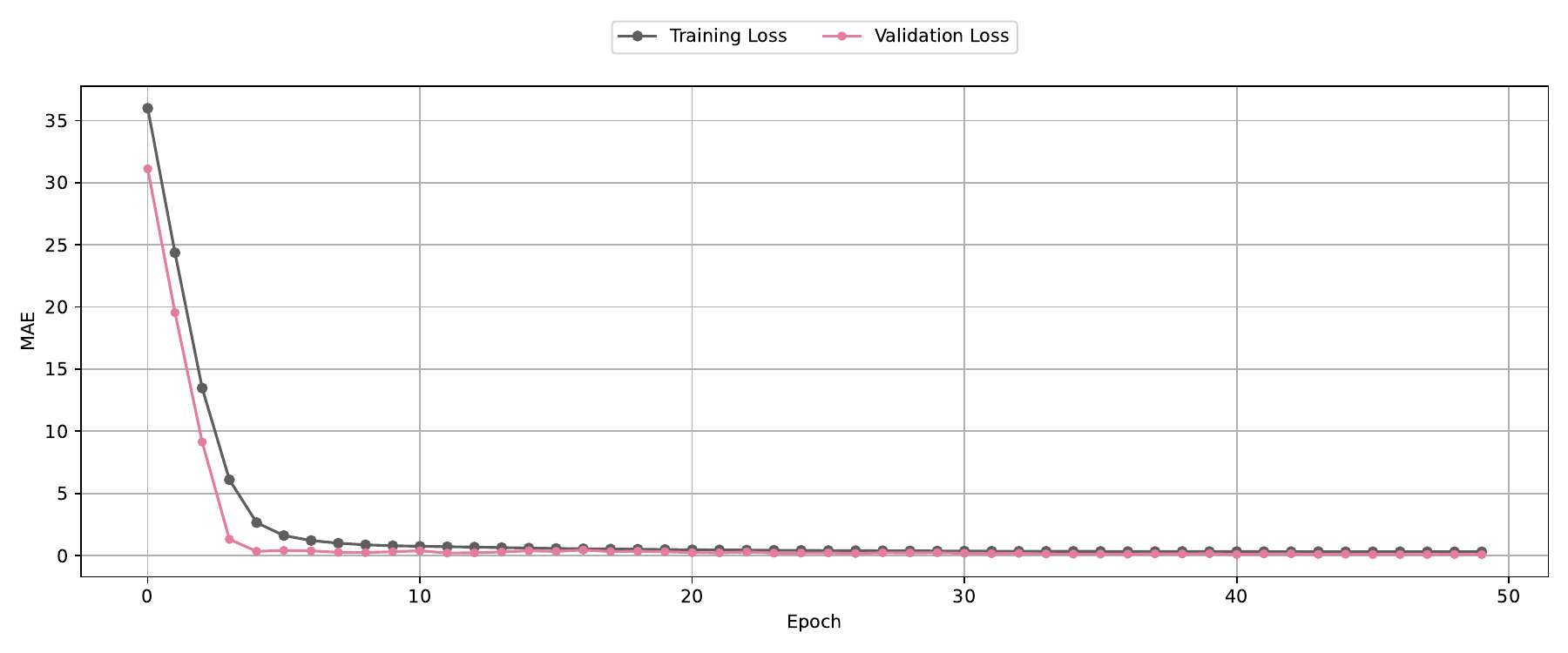}}}\hspace{5pt}
    \subfigure[Household 5]{
        \resizebox*{6.8cm}{!}{\includegraphics{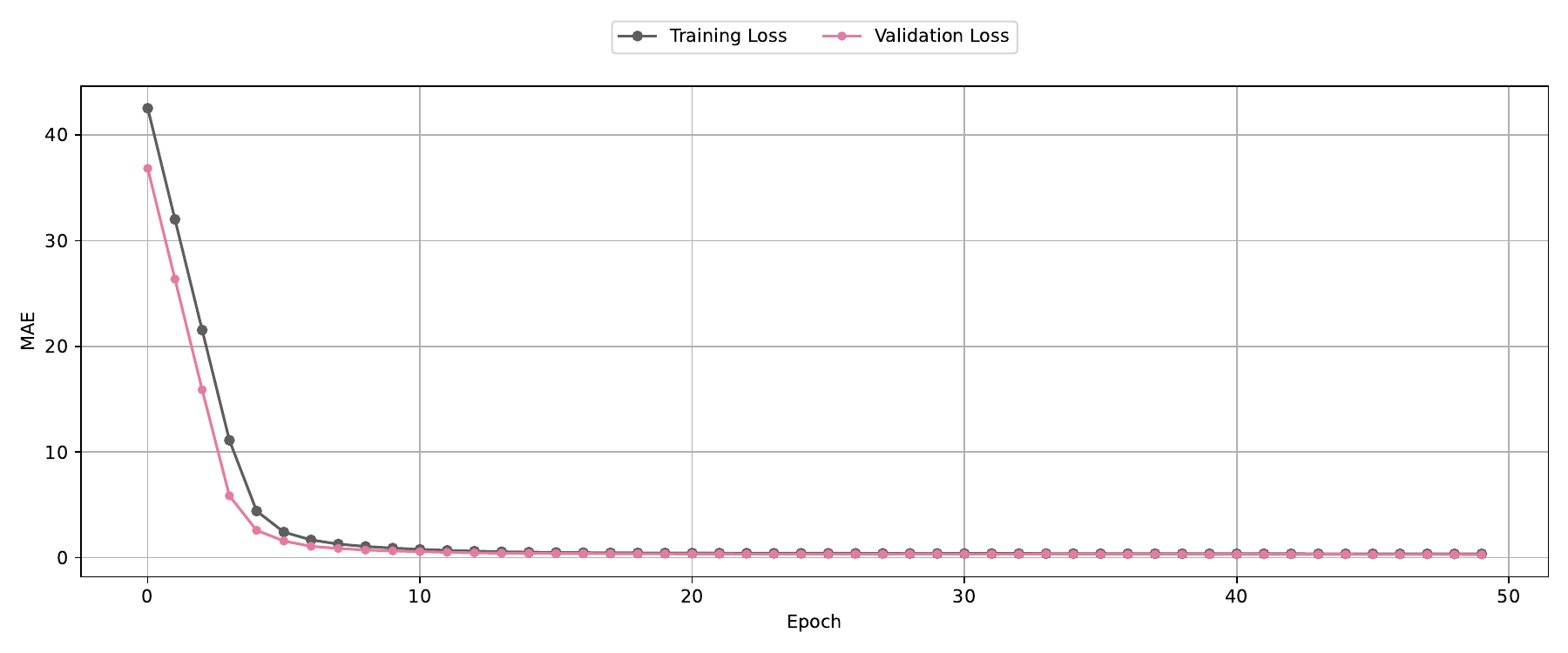}}}
    \caption{AttLSTM learning curves for household 4 and 5}
    \label{fig:Attlosscurves}
\end{figure}

On the other hand, attLSTM performance is relatively close to both models, but generally outperforms the conventional LSTM model in most households, demonstrating lower RMSE and MAPE values. \textcolor{blue}{Given that the household data is not normally distributed, the Wilcoxon Signed-Rank test further supports this observation, indicating a statistically significant difference between LSTM and attLSTM (p=0.0312)}. While attLSTM shows an overall trend of better performance, specific cases like household 1 highlight lower generalizability compared to LGBM. From a generalizability perspective, attLSTM can be better fitting to multiple patterns of household consumption compared to LGBM, which requires hypertuning for each household. \textcolor{blue}{This aligns with the statistical findings, where attLSTM appears similar to LightGBM than to LSTM in capturing certain consumption patterns, where the test suggests no significant difference between attLSTM and LightGBM (p=0.0938)}. In Figure \ref{fig:LSTMlosscurves}  and Figure \ref{fig:Attlosscurves}, the LSTM and attLSTM learning curves are shown respectively for household 4 and household 5, which show the best-performing results.

Regarding training time, Table \ref{tab:runtime} shows that the run time for LGBM (average 3.55 minutes) is significantly lower compared to LSTM (average 9.44 minutes) and attLSTM which is even higher (average 15.11 minutes). This highlights the computational efficiency of LGBM in addition to its good performance. In fact, LSTM and attLSTM models require up to 4× longer training times than LGBM, which in this case can impact scalability in a resource-constrained environments. Overall, in such field where the rapid deployment and low computational overhead are priorities, LGBM provides the best balance of speed and accuracy.

\begin{table}
\centering
\caption{\textcolor{blue}{Total train time (in minutes) for LGBM, LSTM, and attLSTM across all households.}}
\label{tab:runtime}
\resizebox{\textwidth}{!}{%
\begin{tabular}{lccc}
\toprule
\textcolor{blue}{\textbf{Household \#}} & \textcolor{blue}{\textbf{LGBM train time}} & \textcolor{blue}{\textbf{LSTM train time}} & \textcolor{blue}{\textbf{attLSTM train time}} \\
\midrule
\textcolor{blue}{\textbf{Household 1}} & \textcolor{blue}{4.40}   & \textcolor{blue}{10.03} & \textcolor{blue}{15.07} \\
\textcolor{blue}{\textbf{Household 2}} & \textcolor{blue}{3.46}   & \textcolor{blue}{10.05} & \textcolor{blue}{15.9} \\
\textcolor{blue}{\textbf{Household 3}} & \textcolor{blue}{3.35}  & \textcolor{blue}{9.22}  & \textcolor{blue}{15.13} \\
\textcolor{blue}{\textbf{Household 4}} & \textcolor{blue}{3.43}   & \textcolor{blue}{9.22}  & \textcolor{blue}{15.13} \\
\textcolor{blue}{\textbf{Household 5}} & \textcolor{blue}{3.24}   & \textcolor{blue}{8.33}  & \textcolor{blue}{13.45} \\
\textcolor{blue}{\textbf{Household 6}} & \textcolor{blue}{3.44}   & \textcolor{blue}{9.78}  & \textcolor{blue}{15.97} \\
\bottomrule
\end{tabular}%
}
\end{table}

\subsection{Construction of household 4 hot water demand calendar } \label{sec:household4}
To evaluate the performance of the proposed approach on a more granular level, an exhaustive implementation and performance evaluation of our approach is conducted on household 4. Given the conclusions drawn in Section \ref{sec:evaluation}, and considering that the ML model achieved the best performance on household 4, it is evident that household 4 exhibits a clear pattern that can be effectively detected and illustrated using this approach. Therefore, the proposed approach is applied to data of household 4, to identify the shower trends and generate the corresponding hot water demand calendar. 

While the ML model already provides the general hot water demand forecast through $t_\mathrm{mid}$ prediction, the shower event detection using iForest serves to identify specific peak usage patterns within this forecast. This two-step approach enables both broad demand prediction and granular peak usage identification, allowing for more precise operational scheduling of hot water production.

\begin{figure}
\centering
\subfigure[LightGBM performance on household 4 data.]{
\resizebox*{12cm}{!}{\includegraphics{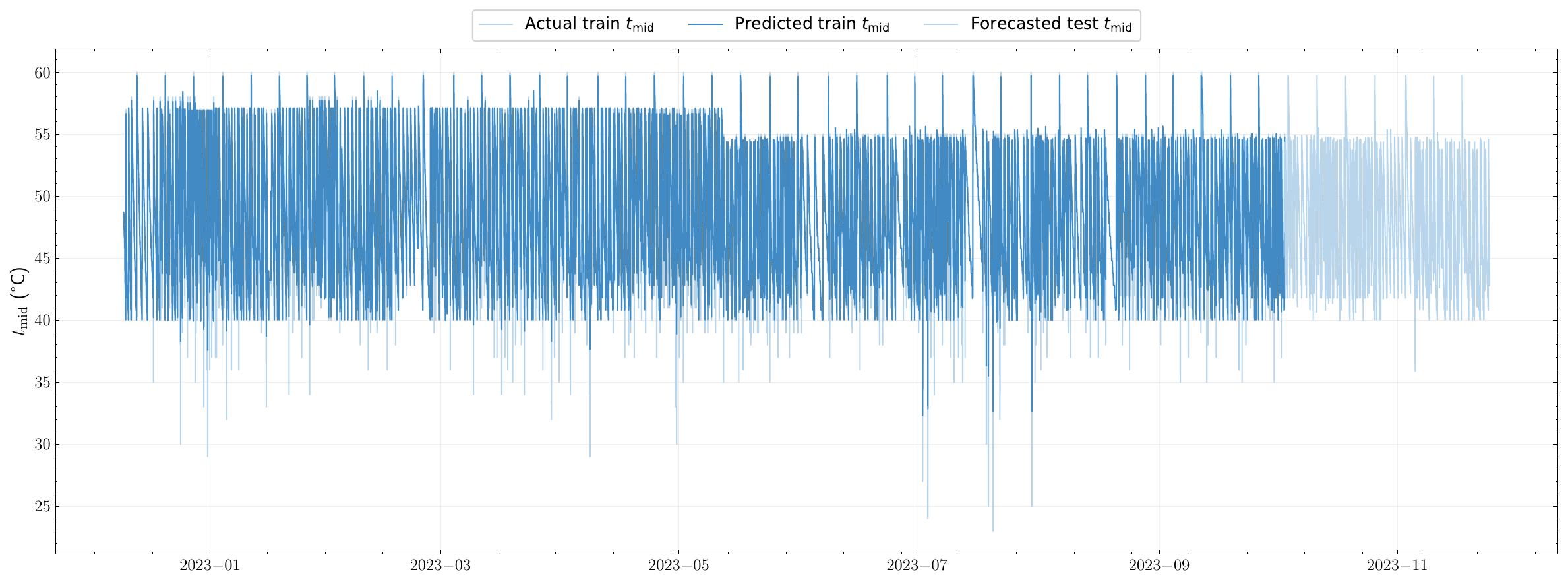}}}\hspace{5pt}
\label{fig:household4forecast}
\subfigure[Closer examination of the detected shower event for household 4.]{
\resizebox*{12cm}{!}{\includegraphics{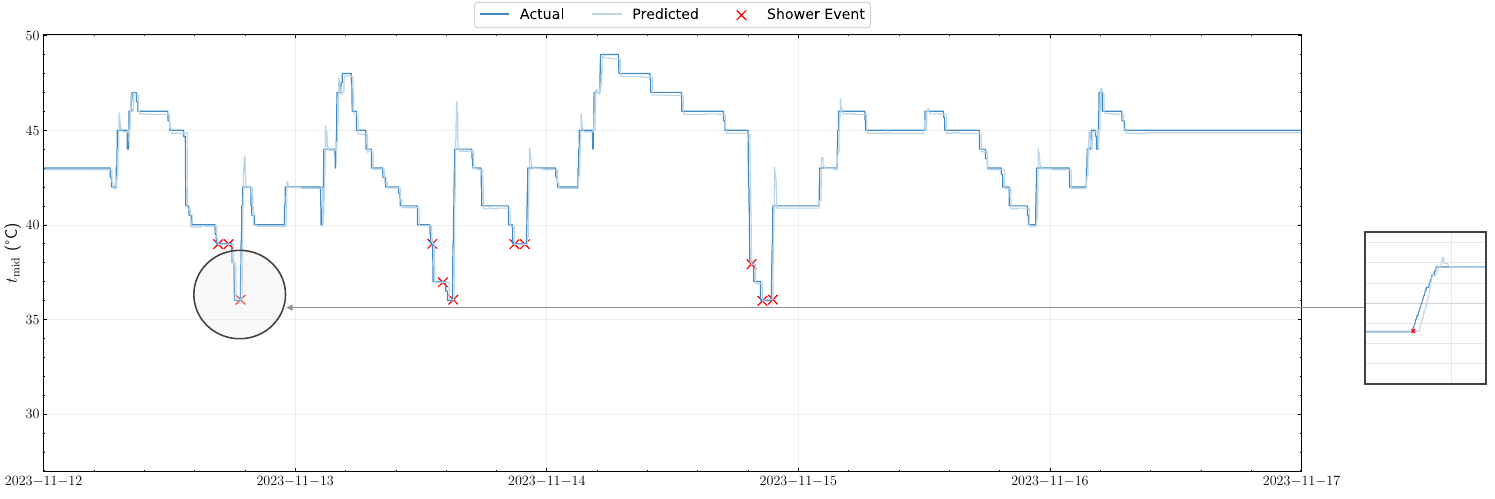}}}
\caption{Detection of shower events using iForest.}
\label{fig:showereventh4}
\end{figure}

Figure \ref{fig:household4forecast} shows the actual and forecasted temperatures $t_\mathrm{mid}$ of household 4 water tank over the period spanning from October 2022 to November 2023. The temperatures range from approximately 25°C to 60°C, with most data points clustering between 40 $^\circ\text{C}$ and 55 $^\circ\text{C}$. Large drops in temperature are observed, which represent historical and predicted trends of shower events. The forecasted time window, highlighted in a lighter blue line, refers to the period between 2023-11-12 and 2023-11-17. LGBM, as the best-performing model among all six household data, is selected to forecast $t_\mathrm{mid}$ temperature for the mentioned time window. 

Once the future consumption of hot water is predicted, the next step is to detect expected shower events during the specified time window. At this stage, iForest is applied to the forecast window data to detect the expected shower events for household 4. The detected shower events are marked in red as shown in Figure \ref{fig:showereventh4}, where the forecast window is zoomed in for clarity. Upon closer examination of one of the highlighted shower events, the HP shows that hot water production begins immediately to compensate for the large and sudden utilization of hot water from the water tank which validates the effectiveness of the approach. The detected shower events, indexed by a time stamp, allow us to build the consumption profile for household 4.

\begin{figure}
    \centering
    \includegraphics[width=1\linewidth]{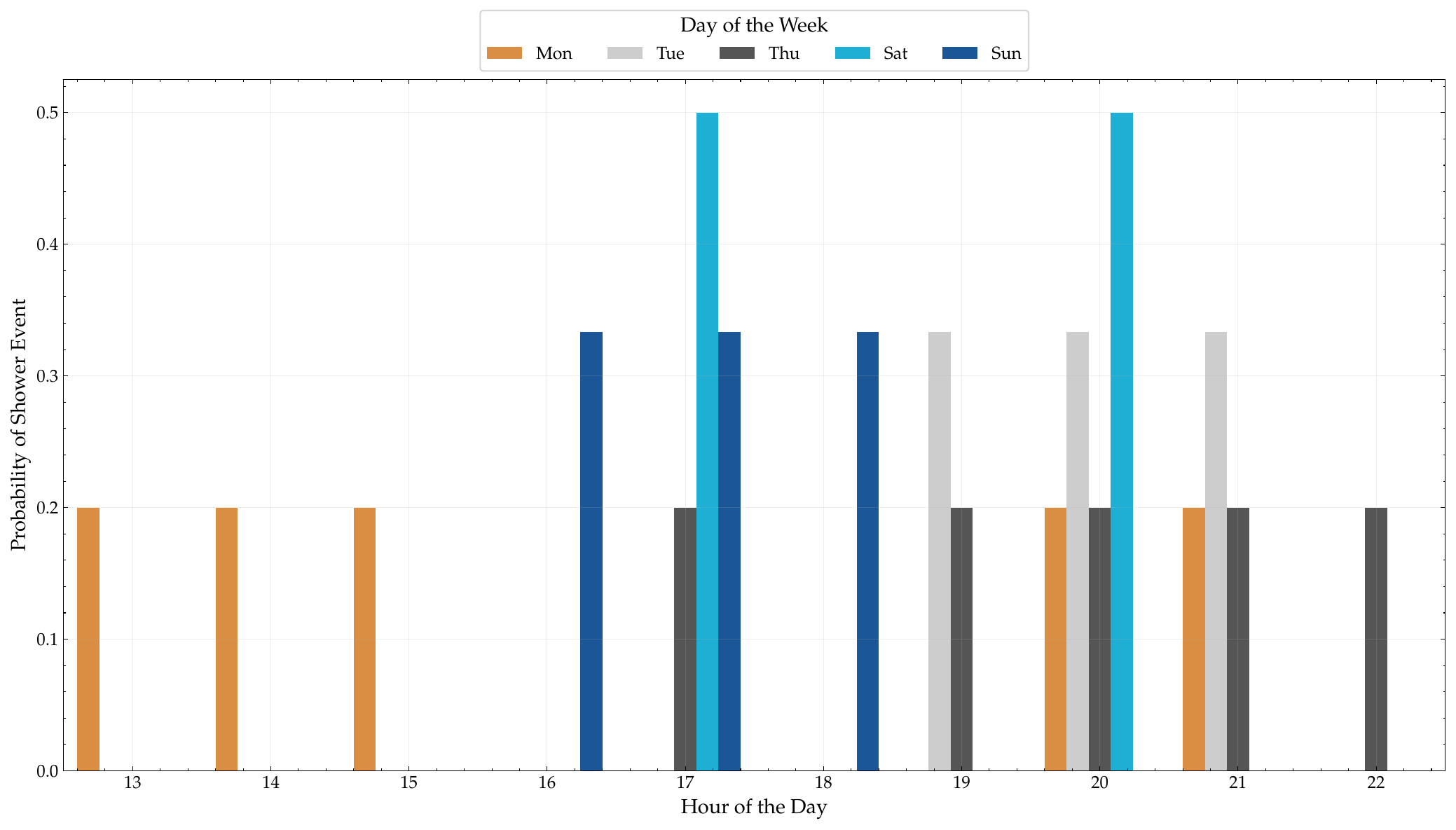}
    \caption{Hourly probability of shower events across weekdays in household 4.}
    \label{fig:calendar}
\end{figure}

In the last step, the calendar representing household 4 demand for hot water is constructed. The date and time of each detected shower event are stored with their corresponding timestamp. From this timestamp, information on the day of the week is extracted, where weekdays are encoded as integers from 0 (Monday) to 6 (Sunday). In addition, the hour is extracted from the timestamp. This allows us to aggregate the frequency of shower events by weekday and hour. This aggregation provides the total count of detected shower events for each hour on a weekday, which is used to build the demand calendar. Upon filtering of shower events by day and hour of the day, the corresponding probability of a shower event is calculated. Figure \ref{fig:calendar} shows the hourly probability of shower events on all days of the week for household 4, with probabilities ranging from 0.0 to 0.5. Notable trends include higher probabilities of showers in the mid to late afternoon, particularly around 15:00 to 17:00, and a general decline toward the evening. Saturday and Sunday exhibit the highest variability, with significant peaks around 14:00 and 15:00, indicating a high probability of a shower event. The absence of data for Wednesday and Friday suggests that shower events are not predicted on those days for the specific time window. As a result, these observed patterns suggest that shower events are more likely to occur in the afternoons and less likely in the late evenings. Thus, it is recommended that hot water production starts in the afternoon, with more attention to Tuesdays and Saturdays. In the final step, the deduced information is transformed into start and stop commands that are communicated with the HP.

\section{Discussion} \label{sec:discussion}

In this paper,  a novel composite approach integrating predictive ML and iForest to forecast the household consumption pattern of hot water is introduced. The approach relies on two key selections of the lightGBM and iForest to predict $t_\mathrm{mid}$ temperature and detect the expected shower events respectively. The selection of iForest in the case of shower event detection enables our approach to dynamically respond to changing behavioral patterns among the different households. This is leveraged by the fact that iForest is a flexible method that does not require selecting a threshold to perform the event detection of interest. As it can be challenging in such a case study to choose a suitable threshold for every household in a real-world setting. Additionally, since data labeling for every household is typically challenging and expensive to perform, and given that the behavior of a household is likely to change between seasons and with time, iForest as an unsupervised anomaly detection method is preferred over other statistical or supervised methods. The results of the experiments conducted demonstrate that the proposed approach achieves the intended goals in terms of forecasting the household demand for hot water. In regard to observed MAPE and RMSE metrics, the integration of lightGBM and iForest shows generally good performance across all household data.

Based on our experimental results, our approach demonstrates several key advantages. The framework shows strong adaptability across different HP types (both GSHP and ASHP) and varying household consumption patterns, as evidenced by its successful implementation across six different households. The use of iForest eliminates the need for household-specific thresholds, making the approach more generalizable. Furthermore, LightGBM showed superior performance compared to LSTM variants, particularly in prediction accuracy, while enabling the generation of household-specific demand calendars that reflect actual usage patterns. \textcolor{blue}{In addition, LGBM shows low computational overhead which suits such an application where training resources are critical.}

While working closely with iForest, a deeper understanding of the best use cases that can benefit from this approach is observed. Specifically, the proposed approach is best suited for households who demonstrate a repetitive pattern in their consumption, otherwise, the resulting consumption calendar will give similar weights to all days and hours of the day. Thus, making it hard to identify useful start and stop information to be pushed to the HP. In this study, the contamination rate for the iForest was preset to 0.05 across all experiments on the household data, but recognizing that the ideal rate might vary among a larger group of households. Despite the generally good performance of the iForest in detecting shower events, the implementation of post-processing was required to refine the detection of shower events. For example, in cases where the heat pump does not start immediately to produce hot water after a temperature drop, the system might falsely detect multiple shower events. Therefore, we filtered out detected events occurring within a 30-minute range to prevent multiple detection of a single event. Consequently, only the most significant drop within a time window was retained as a valid shower event. Additionally, the effectiveness of the iForest is dependent on the accuracy of the preceding ML forecasting stage. Therefore, selecting the best model for each specific case is crucial. This emphasizes the importance of model selection and hyperparameter tuning to ensure reliable identification of the hot water consumption calendar for a household.

\section{Threats to validity}\label{sec:threats}
\subsection*{External Validity}
Given the changing conditions and assumptions, generalizability is the main threat in the construction of approaches for real-world use cases. Generalizability falls under external validity concerns that must be addressed to ensure the reproducibility of the research work. The proposed methods support generalization across multiple components such as in the Data preprocessing pipeline, feature extraction methodology, and model selection and hypertuning, selection of anomaly detection method.

In Figure \ref{fig:framework}, the abstract of the integrated workflow that is designed to be adaptive without any preset specifications for the involved components is presented. Therefore, our approach can be used for other applications in different domains with simple adjustments to the tools or sequence of the proposed tasks. For example, it can be applied to different types of time series data of multivariate and univariate nature.

\subsection*{Limitations and Technical Considerations}
The proposed framework, while demonstrating strong performance in modeling the household hot water demand pattern and including the detection of shower events, exhibits a few limitations that can be categorized into two main areas: model-specific, data dependencies, and dataset considerations.

\subsubsection*{Model-Specific Limitations}
The performance of lightGBM, is particularly sensitive to the variations between the households data patterns. Our implementation of this model, while effective, requires household-specific hyperparameter tuning to ensure the highest prediction performance. In regard to the application of iForest, we selected a contamination rate of 0.05 after rounds of experimentation with multiple different rates. The selected rate yielded the best results with this case study data. However, we did not include these experimental results in the paper to maintain the focus on the primary use case and the implementation of the proposed framework. It important to note that changing this rate can impact the shower detection outcomes produced by iForest.

\subsubsection*{Data Quality Dependencies}
The effectiveness of our framework relies on high-quality data from the top and mid sensors installed within the water tank. To ensure continuous and consistent intervals between recordings, the sensor data are resampled to a 1-minute granularity. This continuity is essential for setting the parameters of the LSTM model.

\subsubsection*{Dataset Considerations}
The dataset used in this study spans the period of 2022-2023, specifically chosen to capture typical household consumption patterns across all four seasons. Evaluating the performance of the proposed approach under unusual conditions, such as lockdown periods that disrupt regular household routines, could be an interesting direction for future research. This is particularly relevant for developing more robust prediction models capable of adapting to sudden behavioral shifts.

A limitation of this study is the relatively small dataset consisting of only six households from Sweden. This limited sample size may affect the generalizability of the models to households located in different climates or with varying water usage patterns. The models developed in this work are based on temperature data and temporal features, which are expected to be applicable to similar scenarios. However, variations in household behavior, regional climates, and cultural differences in hot water usage may impact the performance of the proposed models. Future work could focus on evaluating the robustness of these models by testing them on broader datasets that include households from diverse geographical locations and varying water usage patterns.

\section{\textcolor{blue}{Conclusion and Future Directions}}\label{sec:conclusion}

This research addresses exciting yet understudied challenge of hot water demand forecasting in residential heat pump systems. The work presented in this paper represents an empirical attempt to forecast the consumption of hot water by leveraging ML to learn the behavior of the household inhabitants. Through the development and validation of our proposed approach that combines predictive machine learning with anomaly detection, results demonstrated the superior performance of lightGBM compared to LSTM variants for temperature prediction. Additionally, the integration of iForest for the identification of shower events enabled the generation of household-specific demand calendars across six different household installations. The main characteristic of the proposed approach is that it does not require specifying a threshold for each household, due to the utilization of iForest for detecting shower events. The experimental results show several important insights, including the importance of household-specific model tuning for optimal performance and the key role of temporal feature engineering in improving forecast accuracy.

Building upon these findings, we identify several promising directions for future research. From a technical perspective, the approach could be enhanced through the integration of additional sensor types such as flow rate. Furthermore, methodological extensions could consider the incorporation of weather data and real-time energy pricing to optimize both operational efficiency and cost-effectiveness, ultimately delivering a more economical and user-curated experience. On another level, scaling the approach to larger datasets spanning multiple geographic regions can contribute to a wider representation of household consumption patterns. Based on the insights gained from our experiments, the use of transfer learning to reuse pre-trained models and reduce per-household training can guide future directions of research. Furthermore, the feasibility and generalization of this approach suggest potential applications beyond hot water demand, including general HVAC system optimization and residential energy consumption forecasting, where quantifying energy savings through long-term comparative studies can be investigated.

Future research directions could explore the application of emerging technologies such as foundation models for time series analysis to further enhance the prediction accuracy of hot water demand. However, such investigations would need to carefully balance the potential performance improvements against the practical constraints of real-world heat pump deployments, including computational resources and real-time processing requirements.

\section*{Data availibility }

The datasets generated during and/or analyzed during the current study are not publicly available currently. However,
the authors are willing to provide the data per request.

\section*{Conflict of interest}

The authors have no conflicts of interest to declare that are relevant to the content of this article.

\section*{Acknowledgements} 


\bibliographystyle{tfp}
\bibliography{refs}

\end{document}